\documentclass[lettersize,journal,twoside]{IEEEtran}  

\IEEEoverridecommandlockouts                              

\usepackage{graphics} 
\usepackage{epsfig} 
\usepackage{times} 
\usepackage{amsmath} 
\usepackage{amssymb}  
\usepackage{bbm} 
\usepackage{subfiles} 
\usepackage{blindtext} 
\usepackage{multirow} 
\usepackage{caption} 
\usepackage{tabularx}
\usepackage{subcaption} 
\usepackage{mathtools, nccmath} 
\usepackage{booktabs} 
\usepackage[dvipsnames]{xcolor} 
\usepackage{comment}
\usepackage{placeins} 
\usepackage{bm} 
\usepackage{enumitem} 
\usepackage[linesnumbered,ruled,vlined]{algorithm2e}
\DontPrintSemicolon          
\SetAlgoVlined               
\SetFuncSty{myfuncsty}   
\SetFuncArgSty{myfuncsty} 

\makeatletter
\let\NAT@parse\undefined
\makeatother
\usepackage{hyperref}
\usepackage[table]{xcolor}   
\usepackage{array}
\usepackage{tikz}
\usepackage{scalerel}
\usetikzlibrary{svg.path}
\definecolor{orcidlogocol}{HTML}{A6CE39}
\tikzset{
    orcidlogo/.pic={
        \fill[orcidlogocol] svg{M256,128c0,70.7-57.3,128-128,128C57.3,256,0,198.7,0,128C0,57.3,57.3,0,128,0C198.7,0,256,57.3,256,128z};
        \fill[white] svg{M86.3,186.2H70.9V79.1h15.4v107.1z};
        \fill[white] svg{M108.9,79.1h41.6c39.6,0,57,28.3,57,53.6
            c0,27.5-21.5,53.6-56.8,53.6h-41.8V79.1z
            M124.3,172.4h24.5c34.9,0,42.9-26.5,42.9-39.7
            c0-21.5-13.7-39.7-43.7-39.7h-23.7V172.4z};
        \fill[white] svg{M88.7,56.8c0,5.5-4.5,10.1-10.1,10.1
            c-5.6,0-10.1-4.6-10.1-10.1
            c0-5.6,4.5-10.1,10.1-10.1
            C84.2,46.7,88.7,51.3,88.7,56.8z};
    }
}

\newcommand\orcidicon[1]{\href{https://orcid.org/#1}{\mbox{\scalerel*{
    \begin{tikzpicture}[yscale=-1,transform shape]
        \pic{orcidlogo};
    \end{tikzpicture}
}{|}}}}

\begin{document}
\title{UfM$^*$: Uncertainty from Motion$^*$ for DNN Depth Estimation Using Gaussians}
\author{Soumya Sudhakar,  Sertac Karaman, Vivienne Sze
\thanks{Authors are with the Massachusetts Institute of Technology, Cambridge, MA  02139,  USA.  Emails: {\tt\{soumyas, sertac, sze\}@mit.edu}. This work was funded by the National Science Foundation, Cyber Physical Systems program grant no. 2400541.}}

\maketitle

\begin{abstract} 
Reliable uncertainty estimation is critical for deploying monocular depth deep neural networks (DNNs) in safety-critical robotic systems. Conventional uncertainty methods such as ensembles and sampling-based approaches require multiple inferences per image, incurring substantial compute and memory overhead. 
Moreover, uncertainty predicted from a single image misses out on measuring disagreement between predictions across views of the same region.
We propose Uncertainty from Motion$^*$ (UfM$^*$), an uncertainty estimation algorithm that measures multiview disagreement efficiently by comparing previous and current views using a compact Gaussian mixture, requiring only a single DNN inference per image. Using Gaussians to compute multiview disagreement is not only more compute- and memory-efficient than a prior approach using a point cloud, but also improves uncertainty by measuring disagreement across \textit{regions} of 3D space.
UfM$^*$ paired with aleatoric uncertainty improves expected calibration error by 24-28\% compared to an ensemble, while requiring only 3\% of the energy and 0.02\% of the memory on 100 out-of-distribution ScanNet sequences. 
We demonstrate UfM$^*$ consumes only 63 mJ per 224$\times$224 image while running real-time at 30 FPS on an Arm Cortex-A76 CPU onboard a miniature energy-constrained robot, highlighting that measuring multiview disagreement using Gaussians enables efficient uncertainty for resource-constrained robotic systems.
\end{abstract}

\section{Introduction}
\begin{figure}[t]
\centering
\includegraphics[width=1.0\columnwidth]{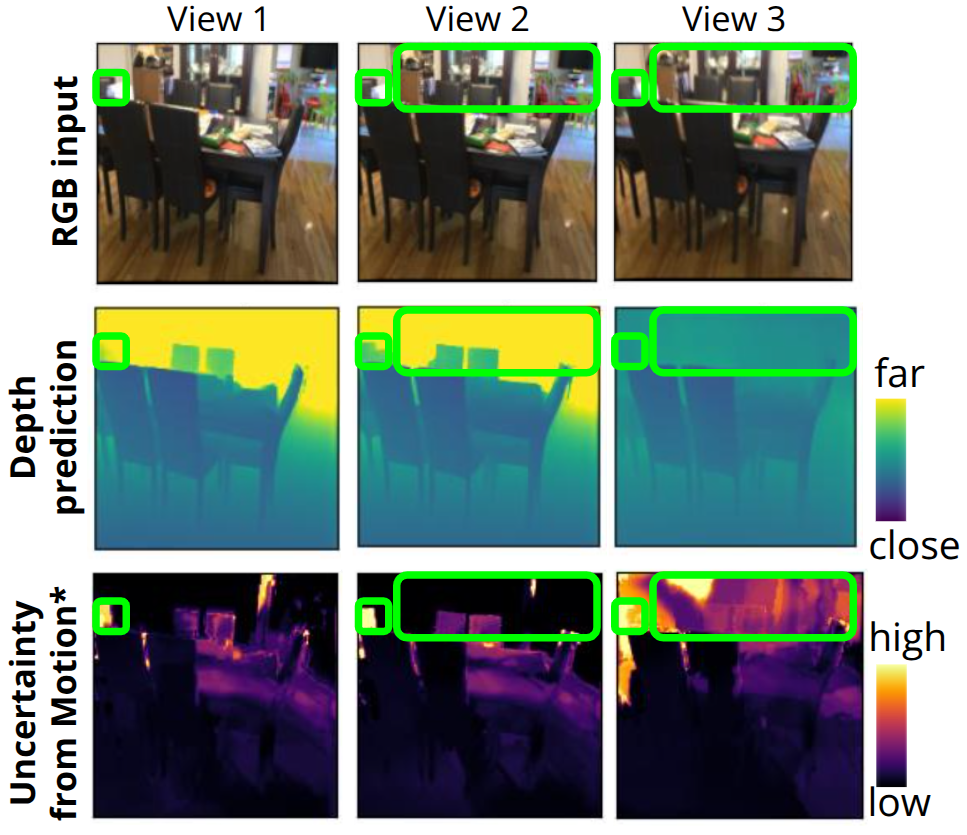} 
\caption{Depth predictions that differ across views of the same 3D space can indicate high uncertainty, as seen by the flickering chair and background (green rectangles) on the depth predictions (middle row) from a Depth Anything V2 model. We propose Uncertainty from Motion$^*$ (UfM$^*$), an algorithm that estimates uncertainty by measuring disagreement across views using Gaussians (bottom row).}
\label{fig:teaser}
\end{figure}
\par For a resource-constrained robot, 
using a monocular depth deep neural network (DNN) to predict depth from images from a single camera enables the robot to understand its environment without large, heavy, and power-hungry depth sensors such as LiDAR and active stereo. However, despite recent successes in training generalizable monocular depth DNNs~\cite{yang2024depth, bhat2023zoedepth}, DNNs can still degrade in accuracy on noisy or out-of-distribution images. For example, when evaluating Depth Anything V2~\cite{yang2024depth} on an indoor ScanNet~\cite{dai2017scannet} dataset, accuracy can drop to zero on some images. Ideally, we would like to detect such failures at runtime in order to take corrective or safety actions~\cite{roemer2023vision, lutjens2019safe, fu2025dectrain}. However, during deployment, we do not have access to ground-truth depth to compute the prediction error; instead, we can estimate DNN uncertainty that ideally correlates to prediction error. 
\par Traditionally, DNN uncertainty estimation methods operate using information from a single image at a time~\cite{lakshminarayanan2017simple, kendall2017uncertainties, amini2020deep, charpentiernatural}, which is useful for tasks operating on random or sparse images. 
In contrast, a monocular depth DNN onboard a robot 
operates on a continuous video stream containing overlapping views of the same 3D scene as seen in the top row of Fig.~\ref{fig:teaser}.
Given that accurate depth predictions should remain geometrically consistent across views of a static scene, measuring multiview disagreement, \textit{i.e.,} when depth predictions are inconsistent across views, can indicate high uncertainty. We see high multiview disagreement in Figure~\ref{fig:teaser}, where the DNN inconsistently predicts the depth of a chair and background across views.
\par In addition to uncertainty quality, using multiple views can lower the computational cost of uncertainty estimation.
Typically, we decompose uncertainty into \emph{aleatoric} (data) uncertainty and \emph{epistemic} (model) uncertainty~\cite{kendall2017uncertainties}. While aleatoric uncertainty is relatively inexpensive~\cite{kendall2017uncertainties}, measuring epistemic uncertainty involves computing disagreement across predictions between different models~\cite{lakshminarayanan2017simple, kendall2017uncertainties, blundell2015weight}. When limited to a single image, we require multiple DNN inferences per image to measure disagreement between different models, making it impractical for resource-constrained robotic platforms~\cite{fu2025dectrain}. Instead, recent works have proposed alternating models over multiple views of the same points in 3D space in order to only need a single inference per image while still measuring disagreement over different models for the tasks of monocular depth estimation and semantic segmentation~\cite{sudhakar2022uncertainty, huang2018efficient}. 
\par While previous works have considered incorporating information from multiple views~\cite{sudhakar2022uncertainty, huang2018efficient, liu2019neural}, they have limitations that can reduce the efficiency and quality of using multiple views. We observe four desirable properties for a multiview method in the resource-constrained setting: 
\begin{itemize}
    \item \textbf{Single inference per image:} Running multiple DNN inferences is too expensive for a resource-constrained robot; instead, we require a method that only requires a single inference per image.
    \item \textbf{Post-hoc uncertainty estimation: }While several methods modify the DNN architecture to take as input multiple views to reduce flickering~\cite{chen2025video, kopf2021robust} or estimate uncertainty~\cite{liu2019neural}, they require expensive retraining, add compute and memory overhead, and suppress information from flickering. Instead, uncertainty estimation that can be applied post-hoc (no architecture modification or retraining) enables application to any monocular depth model and makes flickering of monocular depth predictions useful for uncertainty.
    \item  \textbf{Compact 3D multiview representation:} In order to measure disagreement across views post-hoc, we require a 3D representation of measurements from previous views to compare measurements from our current view. 
    This representation must be compact for real-time deployment on memory- and energy-constrained embedded platforms.
    \item  \textbf{Sensitivity to regional inconsistency:} 
    Multiview disagreement can fail to detect consistently biased predictions, leading to overconfidence. Point-cloud based approaches treat each pixel as independent, though depth predictions are spatially correlated. A representation that checks for regional inconsistency, not simply pointwise, can better avoid some of the overconfident failures. 
\end{itemize}
\par While prior work called Uncertainty from Motion (UfM) proposed by Sudhakar \textit{et al}. satisfies the first two properties~\cite{sudhakar2022uncertainty}, it remains limited due to expensive overhead needed to achieve uncertainty estimation across views. UfM requires a point cloud to represent measurements and uncertainty estimates from previous views, which can become expensive in both compute and memory over time (\textit{e.g.,} millions of points for a room-scale environment). In addition, since a point cloud representation treats each pixel as spatially independent, UfM can result in overconfident uncertainty estimates, and does not improve uncertainty quality compared to ensembles.
\par Our main contribution is a monocular depth DNN uncertainty estimation algorithm called \emph{Uncertainty from Motion$^*$ (UfM$^*$)} that satisfies all the properties above,
leading to improved uncertainty quality, compute efficiency, and memory efficiency. Specifically, we introduce the following: 
\begin{itemize}
    \item \textbf{Gaussian multiview disagreement:} To make measuring multiview disagreement compute- and memory-efficient, we measure disagreement over Gaussian representations of the depth predictions rather than individual point predictions. This shift improves not only efficiency, but also uncertainty quality by mitigating some of the overconfident failure modes.
    \item \textbf{Gaussian correspondences across views:} To enable measuring Gaussian multiview disagreement, 
    we find correspondences between 3D Gaussian components from the current view and from previous views by comparing the overlap of the Gaussian components in the 2D image plane. Importantly, this correspondence method matches Gaussians we \textit{expect} to be consistent, but may not be in 3D space. 
    \item \textbf{Disagreement as distance between Gaussians:} We define the multiview disagreement between corresponding Gaussians to be the \textit{distance} between corresponding Gaussians. 
    \item \textbf{Dense uncertainty from compact Gaussian representation:} In order to get a dense per-pixel uncertainty from the compact Gaussian mixture, we use Gaussian mixture regression (GMR) to compute the multiview disagreement at each 3D point corresponding to each predicted depth pixel.
\end{itemize}
We rigorously evaluate UfM$^*$ and show that it improves calibration relative to an ensemble baseline by $24-28\%$ while requiring only 
$3\%$ of the energy and $0.02\%$ of the memory when paired with aleatoric uncertainty. We further show that UfM$^*$ can act as a stand-alone uncertainty estimator, producing meaningful uncertainty estimates when we apply it to a Depth Anything V2 architecture with no additional aleatoric or epistemic uncertainty estimates. 
\par This paper is organized as follows: Sec.~\ref{sec:related_works} describes relevant research areas and highlights the gaps this work addresses. Sec.~\ref{sec:problem_formulation} describes the problem formulation and notation, Sec.~\ref{sec:multiview} describes multiview disagreement, and Sec.~\ref{sec:algorithm} describes the new algorithm UfM$^*$. Sec.~\ref{sec:metrics} discusses metrics for uncertainty quality, Sec.~\ref{sec:implementation} includes implementation details for the experiment setup, and we rigorously evaluate UfM$^*$ in Sec.~\ref{sec:experimental_results} before concluding in Sec.~\ref{sec:conclusion}.
\section{Related Work}
\label{sec:related_works}
\subsection{Single-view DNN uncertainty estimation}
There has been a long history and recent interest in quantifying uncertainty for DNN predictions for applications such as filtering out high uncertainty predictions in mapping~\cite{liu2019neural} and medical imaging~\cite{Rathore2025EfficientEU}, active learning~\cite{fu2025dectrain, lai2024uncertainty, yang2025active}, and safe motion planning~\cite{roemer2023vision, lutjens2019safe, rabiee2022competence}. Generally, total DNN uncertainty has been divided into aleatoric (data) uncertainty that is irreducible with more training and epistemic (model) uncertainty that is reducible with more training~\cite{kendall2017uncertainties}. We next categorize aleatoric and epistemic uncertainty estimation methods as requiring multiple or single inferences per image. 
\par \textbf{Multiple inferences:} Epistemic uncertainty conventionally has required measuring the disagreement between predictions from multiple models with different weights either via ensembles~\cite{lakshminarayanan2017simple} or sampling-based methods where weights are sampled from a distribution for each inference~\cite{blundell2015weight, gal2017deep, kendall2017uncertainties, franchi2023encoding}. While ensembles have shown state-of-the-art uncertainty quality~\cite{ovadia2019can, postels2022practicality}, they are expensive to run: an ensemble of size $N$ requires $N$ training runs, $N$ copies of DNN weights, and $N$ inferences per image. Recent work has focused on making ensembles more efficient in training~\cite{bousselham2021efficient, Whitaker2022PruneAT, Liu2021DeepEW}, memory~\cite{wen2019batchensemble, laurent2023packed, havasi2020training, baumann2023probabilistic}, and inference~\cite{Li2023TowardsIE, havasi2020training, baumann2023probabilistic}. 
\par \textbf{Single inference:} Aleatoric uncertainty can be estimated by augmenting the network with an additional output head and training under a modified loss to capture irreducible error in the training dataset, an approach introduced in early regression work~\cite{nix1994estimating} and later extended to dense vision tasks~\cite{kendall2017uncertainties}. There has also been a field of work to learn to predict epistemic uncertainty from a single model, either by packing subnetworks into a single network~\cite{laurent2023packed, baumann2023probabilistic}, using knowledge distillation to learn to predict the variance of an ensemble in a single model~\cite{landgraf2024dudes, mariet2021distilling, masakuna2024streamlined}, learning to predict epistemic uncertainty via a modified loss function~\cite{amini2020deep, malinin2018predictive}, using latent features in the bottleneck of a variational autoencoder~\cite{alemi2018uncertainty, charpentiernatural}, or using a distance-aware final layer in the DNN architecture that accounts for distance from the training data~\cite{liu2023simple}. While computationally efficient, these approaches typically rely on modified training objectives and remain less robust under distribution shift compared to ensembles~\cite{postels2022practicality}.
\par Similar to single inference methods, UfM$^*$ also only requires one inference per image. Unlike all single-view methods (multiple inference and single inference), UfM$^*$ does not estimate uncertainty from a single view, but instead takes advantage of multiple overlapping views in order to measure disagreement across views. UfM$^*$ is complementary to single-view methods, and can be added post-hoc to any uncertainty estimation approach; in Sec.~\ref{sec:experimental_results}, we demonstrate UfM$^*$ applied to ensembles~\cite{lakshminarayanan2017simple}, BatchEnsembles~\cite{wen2019batchensemble}, MC-Dropout~\cite{gal2017deep}, evidential~\cite{amini2020deep}, aleatoric~\cite{kendall2017uncertainties}, and alone. 
\par A separate branch of work called conformal prediction does not explicitly estimate epistemic and aleatoric uncertainty separately, but rather provides distribution-free confidence intervals~\cite{angelopoulos2023conformal, fontana2023conformal}. However, conformal methods require held-out calibration data from the deployment distribution and assume exchangeability of the deployment data. In our setting, we do not have access to a calibration set of deployment data, and the data is not exchangeable; UfM$^*$ does not require either to work, but unlike conformal prediction, we do not provide guarantees on the correctness of the uncertainty quality. We show through rigorous empirical evaluation that UfM$^*$ performs well.
\subsection{Multiview scene understanding}
The idea of leveraging multiple views from a video stream to infer 3D structure has a long history in multiview stereo and volumetric fusion literature~\cite{hirschmuller2008stereo, yao2018mvsnet, newcombe2011kinectfusion, whelan2015real}. Specifically for monocular depth DNN methods, recent work has included video-based depth networks that incorporate temporal windows of images to reduce prediction flicker~\cite{chen2025video, luo2020consistent, kopf2021robust}. These methods enforce geometric consistency either by penalizing geometric inconsistency through additional loss terms~\cite{luo2020consistent, chen2025video}, spatio-temporal depth filtering~\cite{kopf2021robust}, or additionally applying post-hoc smoothing~\cite{chen2025video}. While such approaches improve depth accuracy and temporal stability, they require modifications to network architecture or training procedures and aim to suppress geometric inconsistency. In contrast, we do not seek to eliminate geometric inconsistency; instead, we treat disagreement across views as a measurement signal for uncertainty in monocular depth predictions.
\par In parallel, neural rendering methods also use multiple views to construct continuous scene representations from collections of RGB images and poses. While works in this field also quantify uncertainty via a learned aleatoric prediction~\cite{pan2022activenerf}, ensembles~\cite{klasson2024sources}, or frequency of views~\cite{zhao2025ramen}, we consider a different task of estimating a noise model for predictions from a monocular depth DNN and specifically study how disagreement between predictions from multiple views can help estimate uncertainty. 

\subsection{Multiview DNN uncertainty estimation}
Similar to our work, a few prior works also lie at the intersection of DNN uncertainty and geometric consistency across multiple views. Liu \textit{et al.}~\cite{liu2019neural} propose Neural RGB-D, which processes a window of images, constructs a depth probability volume, and treats measurements that agree in 3D space as high confidence. Similar to our work, they leverage geometric inconsistency as an uncertainty signal. However, their approach requires a modified network architecture and retraining, whereas UfM$^*$ operates post-hoc on predictions from any pretrained depth model and can be combined with uncertainty estimation algorithms that differentiate between aleatoric and epistemic uncertainty. 

\par Huang \textit{et al.}~\cite{huang2018efficient} use optical flow to avoid running multiple inferences per image using MC-Dropout and instead estimate disagreement over tracked pixels over multiple images. Similarly, we also use overlap between multiple views to avoid running an ensemble or sampling-based method like MC-Dropout multiple times per image. Unlike Huang \textit{et al.}~\cite{huang2018efficient}, for the task of depth estimation, we require a 3D representation of past measurements in order to compute relative depth for a view from a new camera pose, not just optical flow between pixels. We also avoid using an expensive optical flow DNN that adds to overhead and introduces a secondary DNN noise model. Instead, we use an efficient, albeit noisy reprojection approach, and evaluate the impact of pose noise on the quality of the result. 

\par Closest in spirit to our work, Sudhakar \textit{et al.}~\cite{sudhakar2022uncertainty} introduce Uncertainty from Motion (UfM), which maintains a 3D point cloud to store previous predictions and uncertainty estimates, runs a single inference per image by alternating models, and compares the current depth prediction against the point cloud to estimate multiview disagreement. Similar to UfM, we also alternate models to only run a single inference per image and use a 3D representation to represent previous predictions and uncertainty estimates. Unlike UfM, we do not use a point cloud as the 3D representation since it does not explicitly model spatial correlation in uncertainty, and memory and computational cost scale with the size of the maintained point cloud. For example, in UfM, memory usage can grow substantially (\textit{e.g.}, over 100 MB in room-scale scenes) without aggressive pruning of the point cloud. Adding pruning to UfM reduces memory cost but loses the ability to track or query uncertainty for previously observed regions, hurting uncertainty quality. In contrast, UfM$^*$ uses a compact Gaussian mixture model that captures spatially correlated uncertainty which improves both uncertainty quality and efficiency compared to UfM.

\subsection{Gaussians for scene understanding}
\par There is a rich and growing literature on using Gaussians for representing 3D scenes including applications such as 3D scene reconstruction~\cite{kerbl20233D}, segmentation~\cite{dhawale2020efficient, li2022memory}, mapping~\cite{saarinen20133d, tabib2018manifold, li2022memory, li2024gmmap}, and SLAM~\cite{Matsuki:Murai:etal:CVPR2024, yan2024gs}. To the best of our knowledge,  this work is the first to use a Gaussian mixture model to estimate uncertainty from multiview disagreement for a monocular depth DNN. Our approach builds on key advances from this literature, including efficient projection of 3D Gaussians into the image plane~\cite{kerbl20233D} and segmentation of Gaussian primitives from depth observations~\cite{li2022memory}. In contrast to prior mapping methods that focus on associating measurements that are consistent, our goal is to measure disagreement between measurements (depth predictions) across views, leading us to use a correspondence strategy that matches Gaussians that we expect to be geometrically consistent, but may not be. 
Furthermore, because our objective is uncertainty quality rather than high-fidelity scene reconstruction, we can use more aggressive segmentation to maintain a compact Gaussian representation (\textit{e.g.,} hundreds of Gaussians for a room-scale environment) compared to reconstruction-oriented approaches.
\section{Problem Formulation}
\label{sec:problem_formulation}
In this section, we formulate the monocular depth DNN uncertainty estimation task for a resource-constrained robot operating on a sequence of RGB images. Regarding notation, throughout this article we use uppercase letters for matrices, vectors and scalar constants, lowercase for scalar variables, bolded script for random variables, and calligraphic script for probability distributions, reference frames, sequences, and sets. 
\par Let $(I_{i})_{i\in\mathbb{N}}$ be a sequence of RGB images, and let the pose of the camera at image $i$ be given by rotation $R_i \in SO(3)$ and translation $T_i \in \mathbb{R}^3$. Let $\Theta$ be a set of pretrained monocular depth DNNs, where the $n^{\text{th}}$ DNN $\theta_{n}$ maps image $I_i$ with height $H$ and width $W$ to a dense depth map
\begin{equation}
D_i^{(n)} = \theta_{n}(I_i),
\qquad
D_i^{(n)} \in \mathbb{R}^{H \times W}.
\end{equation}
Let $\Omega \subset \{1,\dots,H\}\times\{1,\dots,W\}$ denote the image pixel domain.
For pixel $u \in \Omega$, the scalar depth prediction is $d_{i,u}^{(n)} := D_i^{(n)}(u)$. Some networks additionally output a learned aleatoric variance
$A_i^{(n)}$~\cite{kendall2017uncertainties}, and certain single model methods networks output a learned epistemic variance $E_i^{(n)}$~\cite{amini2020deep}.
Depending on the base network (ensemble-based, sampling-based, or single model), the set of monocular depth DNNs $\Theta$ may take one of the following forms:
\paragraph{Ensemble-based}
Let $\Theta = \{\theta_1, \dots, \theta_{N}\}$ denote an ensemble of $N$ independently trained depth networks. 
\paragraph{Sampling-based}
Let $\Theta = \{\bm{\theta}_1\}$ denote a single stochastic DNN where each inference uses a stochastic sample of weights; bolded notation since weights of a stochastic DNN are random variables.
\paragraph{Single model}
Let $\Theta = \{\theta_1\}$ denote a single deterministic network. 
\par In all cases, the DNN parameters are fixed and no retraining is performed. For brevity, we drop $n$ from the notation unless specifically mentioned. 
\vspace{0.5em}
\par \noindent
\textbf{Objective.}
We interpret the pixel-wise deterministic depth prediction $d_{i,u}$ as the mean and pixelwise total uncertainty of the depth prediction $v_{i,u}$ as the variance of a Gaussian distribution $\mathcal{N}(d_{i,u},v_{i,u})$. 
 The objective is to estimate $v_{i,u}$ for each pixel incorporating information from multiple views
to obtain a dense uncertainty estimate $V_{i}$ where $V_{i} \in \mathbb{R}^{H \times W}$, while requiring only one inference per image and maintaining low computational and memory overhead suitable for real-time robotic systems. Depending on the base network, the total uncertainty estimate may combine disagreement across views with single-view aleatoric or epistemic terms.
\vspace{0.5em}
\par \noindent
\textbf{Assumptions.}
Camera poses $\{R_i, T_i\}$ are assumed to be given. In experiments, poses are obtained via SLAM~\cite{dai2017bundlefusion, campos2021orb}, and robustness to pose noise is evaluated in
Section~\ref{sec:experimental_results}.
Additionally, we assume a static scene; dynamic objects can violate geometric
consistency across views if not accounted for, and we leave relaxing this assumption for future work.

\section{Multiview Disagreement for Uncertainty}
\label{sec:multiview}
\begin{figure*}[t]
\centering
\includegraphics[width=0.48\textwidth]{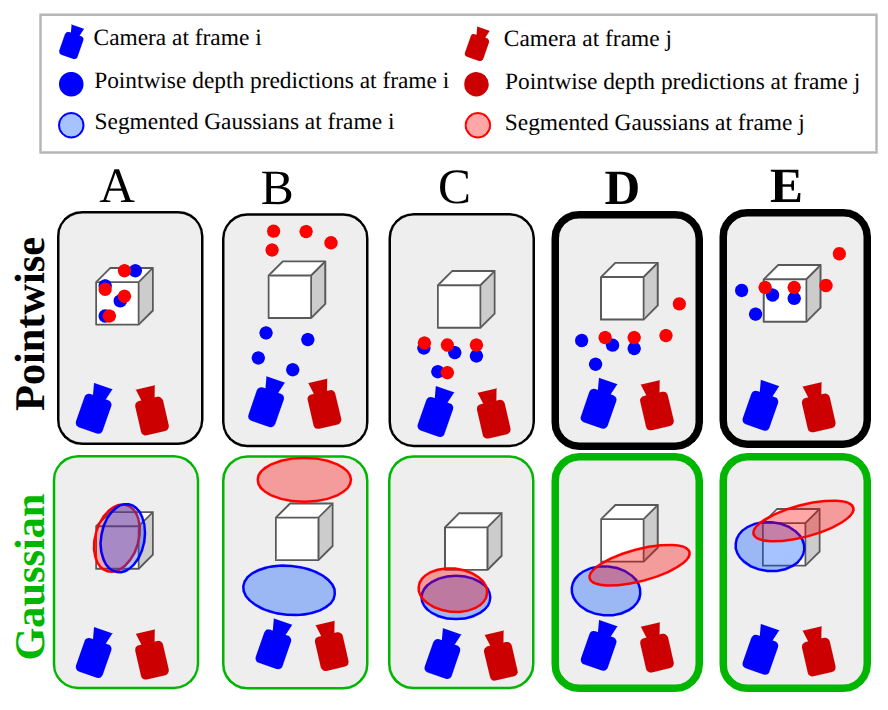}
\includegraphics[width=0.355\textwidth]{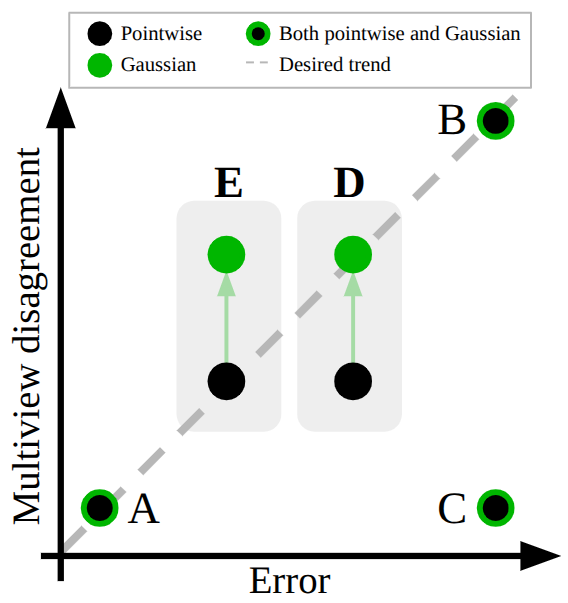}
\caption{
Left: Illustration of use cases with error (distance of measurements from cube object) with pointwise multiview disagreement (distance between red and blue points) and Gaussian multiview disagreement (distance between red and blue Gaussians). Right: Conceptual relationship between depth error and multiview disagreement. 
The desired trend is low error yields low multiview disagreement (case A) and high error yields high multiview disagreement (case B). Both pointwise and Gaussian multiview disagreement can fail when the DNN makes consistently wrong predictions (case C). Pointwise multiview disagreement can produce low disagreement with moderate error when parts of the predictions are consistently wrong while nearby predictions are inconsistent (case D). Using Gaussians for multiview disagreement can better catch these cases since it requires regional consistency, though it can sometimes inflate uncertainty on correct predictions (case E).}
\label{fig:multiview_disagreement}
\end{figure*}
In this section, we formalize the notion of multiview disagreement and motivate its use to measure uncertainty. Intuitively, multiview disagreement measures the degree of geometric inconsistency between depth predictions from multiple views. 
Analogous to how ensemble and sampling-based methods quantify disagreement across \textit{models}, multiview disagreement quantifies disagreement across \textit{views} (by a single model or multiple models). We describe two forms of multiview disagreement: pointwise multiview disagreement which measures disagreement over a point in 3D space described in Sec.~\ref{sec:pointwise_mvd}, and Gaussian multiview disagreement which measures disagreement over a region in 3D space described in Sec.~\ref{sec:gaussian_mvd}. We compare pointwise and Gaussian multiview disagreement and their limitations in Sec.~\ref{sec:gaussian_vs_pointwise_disagreement}. 
\subsection{Pointwise multiview disagreement}\label{sec:pointwise_mvd}
Let $X$ be a 3D occupied point in space that is visible from a set of images with indices $j \in \mathcal{J}$, where the size of the set is $|\mathcal{J}|$.
Let $u_{r}$ be a pixel in image $r \in \mathcal{J}$ that views point $X$ with DNN predicted depth $d_{r,u_{r}}$. Let us assume a perfect correspondence mapping function $u_{j} = \phi_{r\rightarrow j}(u_{r})$ such that $u_{j}$ is a pixel in image $j \in \mathcal{J}$ with depth prediction $d_{j,u_{j}}$ that also views point $X$. The set of depth predictions that view the same point $X$ is given by $\mathcal{D}_{X} = \left\{d_{j,u_{j}}\right\}_{j\in\mathcal{J}}$
and let $\{\widehat{X}_{j}\}_{j\in\mathcal{J}}$ denote the set of 3D point estimates the depth predictions map to in a common reference frame. Then, the sample mean and covariance are
\begin{equation}
\begin{split}
\mu_{\widehat{X}}
& = \frac{1}{|\mathcal{J}|}\sum_{j\in\mathcal{J}} \widehat{X}_{j},\\
\Sigma_{\widehat{X}}
& = \frac{1}{|\mathcal{J}|-1}\sum_{j\in\mathcal{J}}
\left(\widehat{X}_{j}-\mu_{\widehat{X}}\right)
\left(\widehat{X}_{j}-\mu_{\widehat{X}}\right)^{\top}.
\end{split}
\end{equation}
We define \emph{pointwise multiview disagreement} as the trace of the covariance matrix,
\begin{equation}
m_{p}(X) \;=\; \mathrm{tr}\!\left(\Sigma_{\widehat{X}}\right),
\end{equation}
which measures the total 3D spread of the multiview point estimates and has units of meters squared.
\par Under idealized conditions (static scene, perfect correspondence), if all corresponding depth predictions agree, then $m_{p}(X)=0$. Under the same idealized conditions, if there is non-zero pointwise multiview disagreement, then not all predictions in $\mathcal{D}_{X}$ can be simultaneously correct. 
\par In reality, we do not have a perfect correspondence mapping $\phi_{r \rightarrow j}(u_{r})$ and instead have a noisy correspondence mapping $\widetilde{\phi}_{r \rightarrow j}(u_{r})$. For example, prior work used predicted depth reprojection which is reliant on a noisy pose estimate, static world assumption, and predicted depth accuracy~\cite{sudhakar2022uncertainty} such that $\widetilde{\phi}_{r\rightarrow j}(u_{r})$ may not actually be a pixel viewing the same point in 3D space. In addition, depth measurements of dynamic objects may have low error and high multiview disagreement. These violations can break desired behavior, though prior work has shown pointwise multiview disagreement works empirically well~\cite{sudhakar2022uncertainty}. 
\subsection{Gaussian multiview disagreement}\label{sec:gaussian_mvd}
Consider the depth predictions and uncertainty pictured in Fig.~\ref{fig:teaser}, where we see there is geometric structure rather than random noise. This leads us to a main insight of this work: rather than treating each pixel independently and measuring \textit{pointwise} multiview disagreement, we can instead group pixels into spatially coherent regions and measure their \textit{regional} multiview disagreement. We consider Gaussians for our regional primitive as they have been used to compactly and accurately store geometry~\cite{li2024gmmap} and have analytical properties that make the problem tractable. 
\par Specifically, let $\overline{X}$ be a 3D occupied volume in space that is visible from a set of images with indices $j \in \mathcal{J}$, where the size of the set is $|\mathcal{J}|$. Let $\mathcal{U}_{r}$ be a segmented set of pixels in image $r \in \mathcal{J}$ that views region $\overline{X}$, where 
each pixel $u \in \mathcal{U}_{r}$ has predicted depth $d_{r,u}$. 
Let $\mathcal{G}_{\mathcal{U}_{r}}$ denote a 3D Gaussian distribution fitted to the geometry defined by the pixels in $\mathcal{U}_{r}$. Let us assume a perfect correspondence mapping function $\mathcal{U}_{j} = \phi_{r \rightarrow j}(\mathcal{U}_{r})$ where $\mathcal{U}_{j}$ is the set of pixels in image $j\in \mathcal{J}$ that also views region $\overline{X}$ and has corresponding fitted Gaussian distribution $\mathcal{G}_{\mathcal{U}_{j}}$. The set of Gaussians that view the same region across views $\overline{X}$ is given by 
\begin{equation}
    \mathcal{\mathcal{G}}_{\overline{X}} = \left\{\mathcal{G}_{\mathcal{U}_{j}} : j \in \mathcal{J}\right\}.
\end{equation}
Let $\overline{\mathcal{G}}$ be the Wasserstein-Bures barycenter of $\mathcal{\mathcal{G}}_{\overline{X}}$, which is the distribution that minimizes the Wasserstein distance $W_{2}^{2}$ to each Gaussian in $\mathcal{\mathcal{G}}_{\overline{X}}$, 
\begin{equation}
\overline{\mathcal{G}} = \operatorname*{arg\,min}_{\mathcal{G}} \frac{1}{|\mathcal{J}|} \sum_{j \in \mathcal{J}} W_2^{2}(\mathcal{G}, \mathcal{G}_{\mathcal{U}_{j}})
\label{eq:barycenter}
\end{equation}
interpreted as the weighted average distribution~\cite{agueh2011barycenters}. For $|\mathcal{J}| > 2$ Gaussians, the barycenter can be solved using an iterative solution~\cite{agueh2011barycenters} while for $|\mathcal{J}| = 2$ Gaussians, there is an analytic solution. We define \emph{Gaussian multiview disagreement} as
\begin{equation}
m_{g}(\overline{X}) = \frac{1}{|\mathcal{J}|}\sum_{j \in \mathcal{J}} W_{2}^{2} (\overline{\mathcal{G}}, \mathcal{G}_{\mathcal{U}_{j}}),
\label{eq:gmv_exact}
\end{equation}
which is the average Wasserstein-2 squared distance between each Gaussian and the barycenter, also called the Wasserstein variance~\cite{martinet2022variance}, and has units of meters squared.
\par Under idealized conditions (static scene, perfect correspondence, consistent segmentation and fitting), if all corresponding depth predictions agree, then $m_{g}(X)=0$. Under the same idealized conditions, if there is non-zero Gaussian multiview disagreement, it implies at least one depth measurement corresponding to a pixel in $\{\mathcal{U}_{j}\}_{j \in \mathcal{J}}$ is incorrect. As in pointwise multiview disagreement, in reality, we do not have these idealized conditions. In addition, Gaussian multiview disagreement will also fail to detect uncertainty when the DNN predictions are consistently biased and the disagreement magnitude remains bounded by the spread of the predictions. We rigorously evaluate the algorithm we propose based on Gaussian multiview disagreement across diverse datasets in Sec.~\ref{sec:experimental_results}, and show incorporating Gaussian multiview disagreement can improve uncertainty quality compared to prior work based on pointwise multiview disagreement~\cite{sudhakar2022uncertainty}. 
\subsection{Pointwise vs. Gaussian multiview disagreement}\label{sec:gaussian_vs_pointwise_disagreement}
Clearly, using a small set of Gaussians instead of a point cloud is more compact and we will highlight savings in memory and compute in Sec.~\ref{sec:experimental_results}; perhaps a more subtle point is how using Gaussians instead of a point cloud can also improve the uncertainty quality in certain cases. The usefulness of multiview disagreement for improving uncertainty quality depends on whether it tracks prediction error (gray dashed line on right figure in Fig.~\ref{fig:multiview_disagreement}). 
In Fig.~\ref{fig:multiview_disagreement}, we compare and contrast the behavior of pointwise vs. Gaussian multiview disagreement in five different cases. Their behavior diverges in cases where some of the measurements are inconsistently wrong and some measurements are consistently wrong. Since Gaussian multiview disagreement checks whether the Gaussian representation is consistent rather than just each individual point, it can correctly flag a whole region as high disagreement even if some points are consistent. This behavior can reduce overconfidence at the risk of introducing some underconfidence.
\par To summarize, using Gaussian multiview disagreement conservatively estimates uncertainty compared to pointwise multiview disagreement where it can prevent overconfidence in certain scenarios at the expense of sometimes being underconfident. In practice, we will show that the algorithm we propose based on Gaussian multiview disagreement (UfM$^*$) improves overall uncertainty quality compared to previous work based on pointwise multiview disagreement~\cite{sudhakar2022uncertainty}, highlighting the importance of fixing overconfidence. For the remainder of the paper, we refer to Gaussian multiview disagreement as multiview disagreement for brevity.
\section{Uncertainty from Motion$^*$ (UfM$^*$)}
\label{sec:algorithm}
\begin{figure*}[t]
\centering
\includegraphics[width=1.0\textwidth]{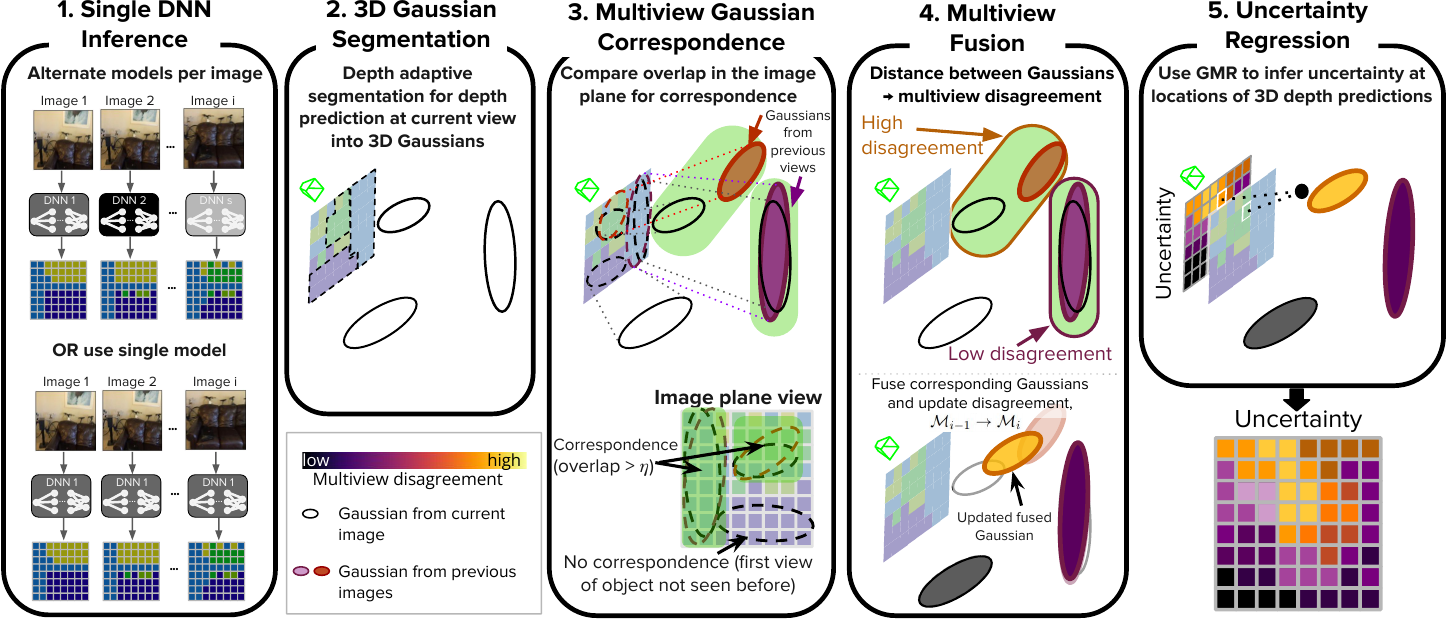} 
\caption{Overview of Uncertainty from Motion$^*$ (UfM$^*$) algorithm.}
\label{fig:alg_overview}
\end{figure*}

We now present our main algorithm, Uncertainty from Motion$^*$ (UfM$^*$). UfM$^*$ involves five steps at each image illustrated in Fig.~\ref{fig:alg_overview} and described below:
\begin{enumerate}
    \item \textbf{Single DNN inference:} choose a model to run a single inference to obtain a depth prediction.
    \item \textbf{3D Gaussian segmentation:} segment the current image into a set of 3D Gaussian components that form a Gaussian mixture model (GMM) that represents the geometry of the depth predictions.
    \item \textbf{Multiview Gaussian correspondence:} establish correspondences between Gaussian components segmented from the current image and Gaussian components from the GMM based on previous views' depth predictions.
    \item \textbf{Multiview fusion:} update the multiview disagreement between corresponding Gaussian components using the squared Wasserstein-2 distance and fuse geometry.
    \item \textbf{Uncertainty regression:} use Gaussian mixture regression (GMR) to compute the multiview disagreement at any 3D point, and query it at pixel predicted depth locations to obtain a dense uncertainty map. 
\end{enumerate}
We next discuss the mixture model used across images, and then discuss each step of the algorithm in detail.
\subsection{Gaussian mixture model for geometry and uncertainty}
GMMs have been used for representing occupancy maps in prior work~\cite{li2024gmmap}, and we extend the representation to compactly measure and store multiview disagreement.
Let $\mathbf{P}$ be the joint random variable 
\begin{equation}
    \mathbf{P} = \begin{bmatrix}
        \mathbf{X} \\
        \mathbf{H} \\ 
    \end{bmatrix}, 
    \mathbf{X} = \begin{bmatrix}
        \mathbf{x} \\
        \mathbf{y} \\
        \mathbf{z} \\
    \end{bmatrix}, 
    \mathbf{H} = \begin{bmatrix}
        \mathbf{m} \\
    \end{bmatrix},
\end{equation}
where $\mathbf{X}$ is the 3D position and $\mathbf{m} \in \mathbb{R}^{+}$ is the estimated multiview disagreement. Note, for methods that produce a dense uncertainty prediction from a single inference (\textit{e.g.,} aleatoric variance $A_{i}$), additional variables may be appended to $\mathbf{H}$ to compactly store the dense uncertainty prediction within the GMM.
At image $i$, let $\mathbf{P}$ be distributed according to an unnormalized Gaussian mixture model $\mathcal{M}_{i}$, as in 
\begin{equation}
    \mathbf{P} \sim \mathcal{M}_{i} = \sum_{k=1}^{K} \pi_{k}\mathcal{N}(\mu_{k},\Sigma_{k}),
\end{equation}
where $\mathcal{N}(\mu_{k},\Sigma_{k})$ is the $k$th Gaussian component of the mixture, $K$ is the number of Gaussians in the mixture, and $\pi_{k}$ is the weight of the $k$th Gaussian of the mixture. 
We refer to the $k$th Gaussian in the mixture at image $i$ as $\mathcal{G}_{k,i} \sim \mathcal{N}(\mu_{k}, \Sigma_{k},  \pi_{k})$ with weight $\pi_{k}$ and mean and covariance
\begin{equation}
    \mu_{k} = \begin{bmatrix}
        \mu_{k,\mathbf{X}} \\
        \mu_{k, \mathbf{H}} \\
    \end{bmatrix},
    \Sigma_{k} = \begin{bmatrix}
        \Sigma_{k,\mathbf{XX}} & \Sigma_{k,\mathbf{XH}}\\
        \Sigma_{k,\mathbf{HX}} & \Sigma_{k,\mathbf{HH}}\\
    \end{bmatrix}
\end{equation}
where we drop the notation for $i$ for the mean, covariance, and weight for brevity.
We assume $\mathbf{m}$ is constant per Gaussian component such that the only nonzero covariance terms are in $\Sigma_{k,\mathbf{XX}}$, which simplifies the algorithm and improves efficiency, while still achieving high uncertainty quality.
\par At image $i$, we have the mixture from previous images $\mathcal{M}_{i-1}$ and the new depth prediction $D_{i}^{(n)}$ from DNN $\theta_{n}$. We next describe the main steps of UfM$^*$ at each image that update the mixture from $\mathcal{M}_{i-1} \rightarrow \mathcal{M}_{i}$, and obtain a dense uncertainty estimate for the depth prediction.
\subsection{Single DNN inference}
We select and run a single DNN for inference, after which we have predicted depth $D_{i}^{(n)}$ and, based on the mode, optionally a predicted aleatoric variance $A_{i}^{(n)}$~\cite{kendall2017uncertainties} and optionally a predicted epistemic variance $E_{i}^{(n)}$~\cite{amini2020deep}. We highlight different modes below:
\subsubsection{Ensemble-based}
Given $\Theta = \{\theta_1, \dots, \theta_N\}$ denotes an ensemble of trained networks (\textit{e.g.,} ensemble~\cite{lakshminarayanan2017simple}, BatchEnsemble~\cite{wen2019batchensemble}), we alternate models by selecting $\theta_{n}$ for a single inference on image $i$ where $n = (i \texttt{ mod } N) + 1$.
\subsubsection{Sampling-based}
Given $\Theta = \{\bm{\theta}_1\}$ denotes a single stochastic DNN (\textit{e.g.,} MC-Dropout~\cite{gal2017deep}), we run a single inference on $\bm{\theta}_1$ which uses a stochastic sample of weights.
\subsubsection{Single model}
Given $\Theta = \{\theta_1\}$ denotes a single deterministic network (\textit{e.g.,} aleatoric only~\cite{kendall2017uncertainties}, evidential~\cite{amini2020deep}), we run a single inference using $\theta_{1}$. 
\par For brevity, we drop $n$ from notation for the remainder of this section. 
\subsection{3D Gaussian segmentation}\label{sec:alg_segmentation}
We first segment the predicted depth $D_{i}$ at image $i$ into a compact mixture as in 
\begin{equation}
    \mathcal{C}_{i} = \sum_{c=1}^{C} \pi_{c}\mathcal{N}(\mu_{c},\Sigma_{c}),
\end{equation}
with $C$ Gaussian components with the $c^{\text{th}}$ Gaussian given by $\mathcal{G}_{c,i} \sim \mathcal{N}(\mu_{c}, \Sigma_{c},  \pi_{c})$ as illustrated by the dashed line ellipsoids in the second panel of Fig.~\ref{fig:alg_overview}. Since grid-based constant segmentation can combine multiple distinct geometric regions into one Gaussian and RGB-adaptive segmentation can either struggle to respect object boundaries at a coarse-grained level~\cite{samet1984quadtree, wei2024gsfusion} or be computationally expensive~\cite{kerbl20233D}, we choose a recent efficient depth-adaptive segmentation method called Single Pass Gaussian Fitting (SPGF)~\cite{li2022memory}.
By adapting Gaussian components using depth discontinuities in the predicted depth, using SPGF reduces the risk of merging distinct surfaces.
Since DNN depth predictions may flicker across images, depth-adaptive segmentation can lead to different segments on sequential images, making finding correspondences more challenging, as discussed in Sec.~\ref{sec:alg_correspondence}.
For details on SPGF, we refer to Li et al.~\cite{li2022memory}; we briefly summarize here. SPGF enables efficient segmentation of a depth map into Gaussians by operating on one pixel at a time; it does so by exploiting spatial correlation between pixels to incrementally construct Gaussians and using depth-adaptive thresholds to detect discontinuities between objects. 
\par After running 3D Gaussian segmentation,
we measure and update multiview disagreement from Gaussians from previous views in the next two steps. 
\subsection{Multiview Gaussian Correspondence}\label{sec:alg_correspondence}
To measure disagreement between Gaussian components across views, we seek a correspondence method that assigns each Gaussian constructed from the current image $\mathcal{C}_i$ either a matching Gaussian from the mixture $\mathcal{M}_{i-1}$ or no correspondence. While a learning-based optical flow approach is an option~\cite{huang2018efficient}, we instead adopt a geometry-based depth reprojection approach to avoid introducing an additional DNN with its own computational overhead and uncertainty model.
We extend prior reprojection-based correspondence methods from points~\cite{sudhakar2022uncertainty, luo2020consistent} to Gaussians by matching individual Gaussian components across views. This treats each component as an independent Gaussian rather than reasoning jointly over the full mixture, which is an approximation, but it yields a practical and efficient correspondence method for our setting.
\par The correspondence procedure consists of two stages: (1) projecting Gaussians from $\mathcal{M}_{i-1}$ and $\mathcal{C}_i$ onto the current image plane, and (2) matching projected Gaussians based on overlap in the image plane, and is illustrated in the third panel in Fig.~\ref{fig:alg_overview}. For brevity, we have dropped reference frame notation for the geometric components of the mixture in other sections, but we introduce here since finding correspondences involves aligning the reference frames. 
\subsubsection{Projection to the Image Plane} Note, $^{\mathcal{R}_{i-1}}\mathcal{M}_{i-1}$ is in the camera reference frame at image $i-1$ while $^{\mathcal{R}_{i}}\mathcal{C}_{i}$ is in the camera reference frame at image $i$. Our goal in this step is to project both of these mixtures onto the image plane reference frame (pixel coordinates) $\mathcal{I}_{i}$ at image $i$. 
\paragraph{Step 1: $^{\mathcal{R}_{i-1}}\mathcal{M}_{i-1} \rightarrow $ $^{\mathcal{R}_{i}}\mathcal{M}_{i-1}$}
We first transform the geometric component of each Gaussian from the previous camera reference frame $\mathcal{R}_{i-1}$ into the current reference frame $\mathcal{R}_i$. Each Gaussian $\mathcal{G}_{k,i-1} \in \mathcal{M}_{i-1}$ is represented by its geometric parameters $
\mathcal{G}_{k,i-1} \sim \mathcal{N}(^{\mathcal{R}_{i-1}}\mu_{k,\mathbf{X}}, ^{\mathcal{R}_{i-1}}\Sigma_{k,\mathbf{XX}}, \pi_k).$
The transformed mean and covariance are:
\begin{align}
^{\mathcal{R}_i}\mu_{k,\mathbf{X}} &= R_{\text{rel}} \, ^{\mathcal{R}_{i-1}}\mu_{k,\mathbf{X}} + T_{\text{rel}}, \\
^{\mathcal{R}_i}\Sigma_{k,\mathbf{XX}} &= 
R_{\text{rel}} \, ^{\mathcal{R}_{i-1}}\Sigma_{k,\mathbf{XX}} R_{\text{rel}}^\top,
\end{align}
where $R_{\text{rel}}$ is the relative rotation and $T_{\text{rel}}$ is the relative translation.
\paragraph{Step 2: $^{\mathcal{R}_{i}}\mathcal{M}_{i-1} \rightarrow $ $^{\mathcal{I}_{i}}\mathcal{M}_{i-1}$} Next, we project the Gaussians to the image plane reference frame $\mathcal{I}_{i}$ at image $i$. The 3D mean is projected to pixel coordinates using
\begin{align}
^{\mathcal{I}_{i}}\mu_{k} = 
\begin{bmatrix}
    f_x \frac{^{\mathcal{R}_i}\mu_{k,x}}{^{\mathcal{R}_i}\mu_{k,z}} + c_x \\
    f_y \frac{^{\mathcal{R}_i}\mu_{k,y}}{^{\mathcal{R}_i}\mu_{k,z}} + c_y
\end{bmatrix}.
\label{eq:proj_to_image_plane_mean}
\end{align}
To approximate the projected covariance, we use the Jacobian for the linear approximation of the transformation~\cite{kerbl20233D} as in 
\begin{equation}
A =\begin{bmatrix}
\frac{f_x}{\mu_z} & 0 & -\frac{f_x \mu_x}{\mu_z^2} \\
0 & \frac{f_y}{\mu_z} & -\frac{f_y \mu_y}{\mu_z^2}
\end{bmatrix},
\end{equation}
evaluated at $^{\mathcal{R}_i}\mu_{k,\mathbf{X}}$. The projected covariance is
\begin{equation}
^{\mathcal{I}_i}\Sigma_k = A \, ^{\mathcal{R}_i}\Sigma_{k,\mathbf{XX}} A^\top,
\label{eq:proj_to_image_plane_covar}
\end{equation}
and the resulting 2D Gaussian is
\begin{equation}
^{\mathcal{I}_i}\mathcal{G}_{k,i-1} \sim \mathcal{N} \left(^{\mathcal{I}_i}\mu_k,^{\mathcal{I}_i}\Sigma_k,\pi_k\right).
\end{equation}
We consider only Gaussians whose projected mean lies inside image bounds as candidates for correspondences.
\paragraph{Step 3: $^{\mathcal{R}_{i}}\mathcal{C}_{i} \rightarrow $ $^{\mathcal{I}_{i}}\mathcal{C}_{i}$} We repeat the process for projecting 3D Gaussians from the current image to the image plane $\mathcal{G}_{c,i} \in \mathcal{C}_i$ using Eq.~\ref{eq:proj_to_image_plane_mean}-\ref{eq:proj_to_image_plane_covar}, yielding
\[
^{\mathcal{I}_i}\mathcal{G}_{c,i}
\sim
\mathcal{N}
\left(
^{\mathcal{I}_i}\mu_{c},
^{\mathcal{I}_i}\Sigma_{c},
\pi_{c}
\right).
\]

\subsubsection{Gaussian matching in the image plane}
Given we now have projected Gaussians from $^{\mathcal{I}_i}\mathcal{M}_{i-1}$ and $^{\mathcal{I}_i}\mathcal{C}_i$ in the image plane, we can establish correspondences between Gaussians. The goal of establishing a correspondence is identifying which Gaussians \textit{should} match in their measurements, not necessarily which Gaussians actually match in their measurements so that disagreement remains measurable. For intuition, consider how in Fig.~\ref{fig:teaser}, we compared patches in the image plane that overlapped (green rectangles) and observed disagreement in 3D (depth predictions). 
Similarly, to identify which Gaussians should match in their measurements, we mark as correspondences projected Gaussians from $^{\mathcal{I}_i}\mathcal{M}_{i-1}$ and $^{\mathcal{I}_i}\mathcal{C}_i$ that have high overlap in the image plane; in Sec.~\ref{sec:alg_fusion}, we will measure the distance of the corresponding 3D Gaussians to measure their disagreement. 
\par Specifically, for each projected Gaussian $^{\mathcal{I}_i}\mathcal{G}_{c,i}$ from the current image, we query nearby projected Gaussians from $\mathcal{M}_{i-1}$ using an R-tree~\cite{guttman1984r} to obtain a candidate set 
$^{\mathcal{I}_i}\mathcal{G}_{c,i} \leftrightarrow \{^{\mathcal{I}_i}\mathcal{G}_{k',i-1}, \dots \}_{k' \in near(^{\mathcal{I}_i}\mathcal{G}_{c,i})}$.
For each candidate pair, to compute the overlap in the image plane, we choose to compute the Bhattacharyya coefficient which quantifies the overlap between the two distributions from zero to one, as in
\begin{equation}
    \begin{split}
     \beta_{c,k'} &= \exp\left(-\frac{1}{8}(^{\mathcal{I}_i}\mu_{k'} - ^{\mathcal{I}_i}\mu_{c})^\top\left(\frac{1}{2}(^{\mathcal{I}_i}\Sigma_{k'} + ^{\mathcal{I}_i}\Sigma_{c})\right)^{-1}\right.\\
    & (^{\mathcal{I}_i}\mu_{k'} - ^{\mathcal{I}_i}\mu_{c}) \left.- \frac{1}{2} \ln \frac{\left|\frac{1}{2}(^{\mathcal{I}_i}\Sigma_{k'} + ^{\mathcal{I}_i}\Sigma_{c})\right|}{\sqrt{|^{\mathcal{I}_i}\Sigma_{k'}| |^{\mathcal{I}_i}\Sigma_{c}|}}\right),
    \end{split}
\end{equation}
where $\beta_{c,k'} := \beta (^{\mathcal{I}_i}\mathcal{G}_{c,i}, ^{\mathcal{I}_i}\mathcal{G}_{k',i-1})$. 
We define the correspondence set
\[
\mathcal{F}_{c,i}
=
\left\{
^{\mathcal{I}_i}\mathcal{G}_{k',i-1} | k'\in \text{near}(^{\mathcal{I}_i}\mathcal{G}_{c,i}) \; \land \; \beta_{c,k'} \ge \eta\right\},
\]
where a pair is kept as a correspondence if the overlap is over a threshold hyperparameter $\eta$.
When multiple Gaussians from previous images overlap each other in the image plane ($|\mathcal{F}_{c,i}| > 1$), a closer surface can occlude a farther one, such that the farther away surface may not be in view. To reduce incorrect correspondences in these cases, we filter out a Gaussian with mean farther away than a closer Gaussian if their overlap in the image plane exceeds a threshold.
\par Finally, we obtain a set of image-plane correspondences
$^{\mathcal{I}_{i}}\mathcal{G}_{c,i} \leftrightarrow \{^{\mathcal{I}_{i}}\mathcal{G}_{k,i-1}\}$ (or $\varnothing$).
Each projected Gaussian $^{\mathcal{I}_{i}}\mathcal{G}_{\cdot,i}$ is associated with its originating 3D Gaussian
$^{\mathcal{R}_{i}}\mathcal{G}_{\cdot,i}$ and so we obtain the final correspondence for each Gaussian as in
\[
^{\mathcal{R}_{i}}\mathcal{G}_{c,i} \leftrightarrow \{^{\mathcal{R}_{i}}\mathcal{G}_{k,i-1}, \dots \}
\quad \text{or} \quad
^{\mathcal{R}_{i}}\mathcal{G}_{c,i} \leftrightarrow \varnothing.
\]
Gaussian correspondences are not 1:1 to account for segmentation changes across views with different depth predictions for the same surface. For example, a chair may be segmented into two Gaussians in one view and one Gaussian in the next.
\begin{figure*}[t]
\centering
\includegraphics[width=1.0\textwidth]{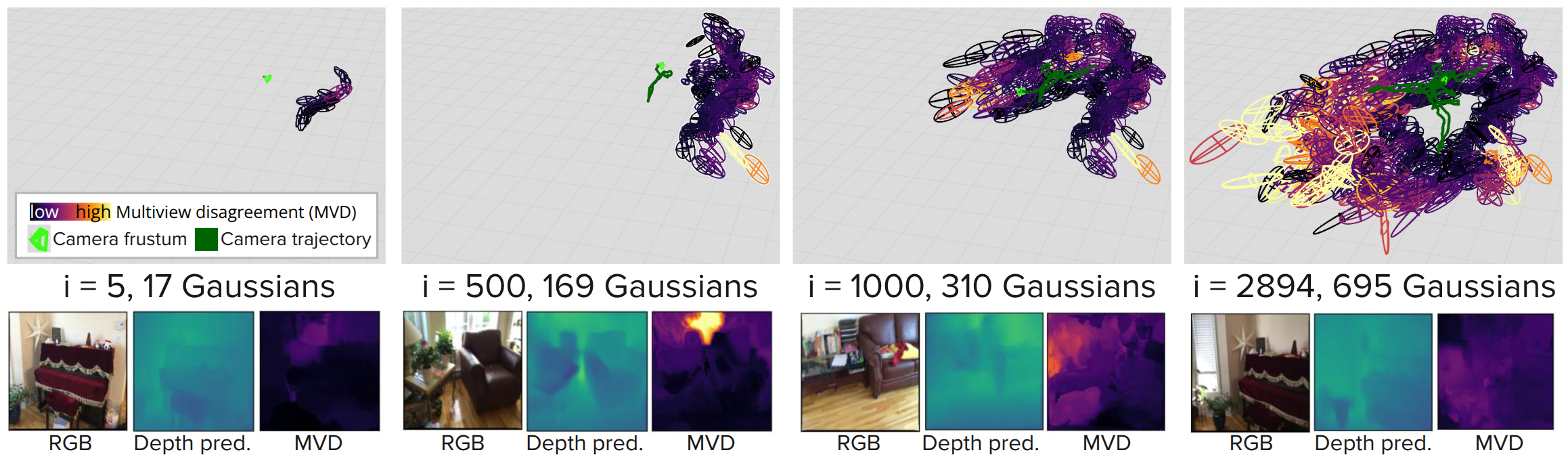} 
\caption{UfM$^*$ constructs a GMM from noisy DNN depth predictions to calculate multiview disagreement (represented by color of the Gaussians) for a room-scale environment. We see low multiview disagreement on the right side of the room and high multiview disagreement on the left side of the room. The representation is compact, requiring only 695 Gaussians.
}
\label{fig:gmm_in_progress}
\end{figure*}

\subsection{Multiview Fusion}
\label{sec:alg_fusion}
Given the established correspondences between Gaussians in $^{\mathcal{R}_{i}}\mathcal{C}_i$ and $^{\mathcal{R}_{i}}\mathcal{M}_{i-1}$, we now measure the multiview disagreement between the 3D corresponding Gaussians and update the global mixture to obtain $^{\mathcal{R}_{i}}\mathcal{M}_i$, as illustrated in the fourth panel in Fig.~\ref{fig:alg_overview}. Since all operations are now in the current camera reference frame $\mathcal{R}_{i}$, 
we drop the reference frame notation for brevity. 
\par For each Gaussian with no correspondence $\mathcal{G}_{c,i} \leftrightarrow \varnothing$, we treat this as the first observation of a new region in 3D, and the corresponding 3D Gaussian is directly added to the mixture:
\[
\mathcal{M}_i \leftarrow \mathcal{M}_{i-1} \cup \mathcal{G}_{c,i}.
\]
If $\kappa$ such Gaussians are added,
the total number of mixture components is updated as $K \leftarrow K + \kappa$.
\par For each pair of corresponding Gaussians in $\mathcal{G}_{c,i} \leftrightarrow \{\mathcal{G}_{k,i-1}, \dots \}$, we merge their geometric components and update the multiview disagreement based on their distance. Instead of solving for the barycenter of more than two Gaussians which requires an iterative solution~\cite{agueh2011barycenters}, we approximate by computing the analytic solution for the barycenter for two Gaussians at a time which lies on the optimal transport geodesic, and repeat if there is more than one corresponding Gaussian. Note, this results in order-dependent behavior. Specifically, we update the Gaussian by moving a weighted $\lambda$ of the distance along the optimal transport geodesic toward the new Gaussian. The weight
\[
\lambda = \frac{\pi_{c}}{\pi_k + \pi_{c}},
\]
reflects the relative weights of each Gaussian. Note, $\pi_{c}$ can be weighted by the correspondence overlap to reflect trust in the correspondence match. 
The mean is updated as
\[\mu_{k,\mathbf{X}} \leftarrow (1-\lambda)\mu_{k,\mathbf{X}} + \lambda \mu_{c,\mathbf{X}},
\]
while the covariance is updated using the optimal transport geodesic between two Gaussians. Specifically, we define
\[
Q = \Sigma_{k,\mathbf{XX}}^{1/2}
      \Sigma_{c,\mathbf{XX}}
      \Sigma_{k,\mathbf{XX}}^{1/2}, 
\]
The covariance is updated using the optimal transport geodesic:
\[\Sigma_{k,\mathbf{XX}} \leftarrow \left((1-\lambda)I + \lambda Z\right)\Sigma_{k,\mathbf{XX}}\left((1-\lambda)I + \lambda Z
\right)^\top,\]
where
\[Z =\Sigma_{k,\mathbf{XX}}^{-1/2}Q^{1/2}\Sigma_{k,\mathbf{XX}}^{-1/2}.\]
The squared 2-Wasserstein distance between the Gaussians is
\begin{equation}
W_2^2(G_{k,\mathbf{X}}, G_{c,\mathbf{X}})=\|\mu_{c,\mathbf{X}} - \mu_{k,\mathbf{X}}\|_2^2+\operatorname{tr}\left(\Sigma_{k,\mathbf{XX}}+\Sigma_{c,\mathbf{XX}}-2 Q^{1/2}\right).
\end{equation}
We interpret the $W_2^2(G_{k,\mathbf{X}}, G_{c,\mathbf{X}})$ as the new measurement on the multiview disagreement $m_{k}$, and update $m_{k}$ for the Gaussian via a running average as in
\begin{align}
m_k &\leftarrow (1-\lambda)m_k + \lambda W_2^2(G_{k,\mathbf{X}}, G_{c,\mathbf{X}}).
\end{align}
Finally, the mixture weight is accumulated using $\pi_k \leftarrow \pi_k + \pi_{c}$. We finish the process for updating $\mathcal{M}_{i-1} \rightarrow \mathcal{M}_{i}$ after adding all Gaussians without correspondences and fusing all Gaussians with correspondences. We show a visual demonstration of $\mathcal{M}_{i}$ at various points during a ScanNet sequence in Fig.~\ref{fig:gmm_in_progress}. We visualize $m_k$ by the color of the Gaussians; we can see that this method is able to spatially capture multiview disagreement compactly, using less than 700 Gaussians for a room-sized environment. 
\subsection{Uncertainty Regression} \label{sec:gmr}
After fusion, 
we now compute a dense uncertainty estimate for each pixel via Gaussian mixture regression (GMR)~\cite{sung2004gaussian}, as illustrated in the last panel in Fig.~\ref{fig:alg_overview}. We briefly summarize the GMR process. To allow querying uncertainty at arbitrary 3D locations, we introduce a weak prior with hyperparameters $\mu_{0}$ and $\Sigma_{0}$:
\[\mathcal{Q}_{\mathbf{H}|\mathbf{X}}(h|x) = \mathcal{N}(\mu_0, \Sigma_0).\]
For each pixel $(u,v)$ with predicted depth $d_{i,u}$, we compute its 3D location 
\[\mathbf{x} =\begin{bmatrix}\frac{u-c_x}{f_x} d_{i,u} \\
\frac{v-c_y}{f_y} d_{i,u} \\
d_{i,u}
\end{bmatrix}.\]
Conditioning the mixture on $\mathbf{X}=\mathbf{x}$ gives
\[
p(\mathbf{H}|\mathbf{X}=\mathbf{x}) = \sum_{k=0}^{K} w_k(\mathbf{x}) \mathcal{N}\big(\mathbf{H} \mid m_k(\mathbf{x}),
\Sigma_{\mathbf{H}|\mathbf{X},k}\big).\]

The weights are
\[w_k(\mathbf{x})=\frac{\pi_k \mathcal{N}(\mathbf{x}|\mu_{k,\mathbf{X}}, \Sigma_{k,\mathbf{XX}})}{\sum_{l=1}^{K}\pi_{l} \mathcal{N}(\mathbf{x}|\mu_{l,\mathbf{X}}, \Sigma_{l,\mathbf{XX}})+ \pi_0}.\]

Since the cross-covariance blocks $\Sigma_{k\mathbf{HX}}$ are zero, the conditional mean simplifies to
\[\mathbb{E}[\mathbf{H}|\mathbf{X}=\mathbf{x}] = \sum_{k=0}^{K}w_k(\mathbf{x}) \mu_{k,\mathbf{H}}=
\begin{bmatrix}
\hat{m}(\mathbf{x})
\end{bmatrix}.
\]
\par After using GMR to query $\hat{m}(X)$ for each 3D location of the depth prediction, we obtain a dense multiview disagreement map $\widehat{M}(X)$. As an optional step, since the depth-adaptive segmentation in Sec.~\ref{sec:alg_segmentation} can lead to instability in the uncertainty since it is based on a potentially flickering DNN predicted depth, we can add an exponential moving average to the dense multiview disagreement map $\widehat{M}_{i}(X) \leftarrow (1-\alpha)\widehat{M}_{i-1}(X) + \alpha\widehat{M}_{i}(X)$, where $\alpha$ is a hyperparameter. 
\par To obtain the total uncertainty estimate $V_i$, we may use $\widehat{M}(X)$ alone, or add it to the single-view learned aleatoric variance $A_i$ or single-view learned epistemic variance $E_i$ when available. The three uncertainty estimates are not necessarily mutually exclusive; disentanglement of multiview disagreement is non-trivial since it can arise due to aleatoric sources (\textit{e.g.,} changes in lighting), epistemic sources (\textit{e.g.,} varying the model), and additional information from multiple views (\textit{e.g.,} zooming out of a close-up view of a photo).
Nevertheless, we sum $\widehat{M}(X)$ because it provides an additional signal based on geometric inconsistency across views, often increasing uncertainty in regions where single-view $A_i$ and $E_i$ remain low. Unlike a global rescaling of $A_i$ or $E_i$, this selectively increases uncertainty where it is detected to be inconsistent across views without uniformly making predictions more conservative. We evaluate this effect empirically in Sec.~\ref{sec:experimental_results}.
\section{Metrics for Uncertainty}
\label{sec:metrics}
Although uncertainty does not have ground truth, we expect high quality uncertainty to correlate with prediction error: uncertainty should increase when predictions are inaccurate and decrease when predictions are accurate. One metric that measures this relationship is negative log-likelihood (NLL), which measures how well the predicted distribution explains the observed depth. For a Gaussian predictive distribution, the per-pixel NLL is
\begin{equation}
    NLL = \frac{1}{2v_{i,u}}(\overline{d}_{i,u}-d_{i,u})^2 + \frac{1}{2}\ln(2\pi v_{i,u}),
\end{equation}
where $\overline{d}_{i,u}$ is the ground-truth depth. While NLL is a proper scoring rule, it entangles uncertainty quality with prediction accuracy such that improvements in prediction accuracy alone can reduce NLL even if the uncertainty estimate does not improve. Consequently, NLL is most informative when comparing methods with similar depth predictions. 
\par Beyond NLL, calibration curves are commonly used to assess whether predicted confidence matches empirical accuracy and to diagnose systematic overconfident or underconfident uncertainty estimates~\cite{guo2017calibration, amini2020deep, sudhakar2022uncertainty}. To construct a $\delta$ calibration curve, we first compute per-pixel (1) confidence defined as the probability mass of the predicted distribution $\mathcal{N}(d_{i,u}, v_{i,u})$ over the interval $[(1-\delta)d_{i,u}, (1+\delta)d_{i,u}]$ where $\delta = 0.25$, and (2) the corresponding accuracy indicator which is one if the ground-truth depth falls inside the same interval and zero if it falls outside. 
We bin pixels by confidence into $B$ bins and plot $conf(b)$ (average confidence of pixels in bin $b$) and $acc(b)$ (average accuracy of pixels in bin $b$) to produce a calibration curve. Under perfect calibration, confidence of each bin $conf(b)$ should equal accuracy of each bin $acc(b)$. Deviations from this behavior are summarized by the expected calibration error ($ECE_{\delta}$),
\begin{equation}
    ECE_{\delta} = \sum^{B}_b \frac{|b|}{S}\left| acc(b)-conf(b)\right|,
\end{equation}
where $|b|$ denotes the number of pixels in confidence bin $b$ and $S$ is the total number of pixels in the sequence. Note, this metric relies on an interval that depends on the predicted depth, making bins less directly comparable across methods with different depth predictions.
\par We additionally evaluate the quantile calibration curve used in prior work~\cite{kuleshov2018accurate}. For each quantile level $q_{b}$, we compute the predicted quantile which is the inverse CDF of $\mathcal{N}(d_{i,u}, v_{i,u})$ for each pixel. For each $q_{b}$, the corresponding observed frequency $obs(q_{b})$ is the fraction of pixels with ground-truth depth at or below the predicted quantile. Under perfect calibration, the quantile level $q_{b}$ should equal the observed frequency $obs(q_{b})$ for each bin $b$. Deviations from this behavior are summarized by the expected calibration error ($ECE_{q}$),
\begin{equation}
    ECE_{q}= \frac{1}{B}\sum^{B}_b\left| obs(q_{b})-q_{b}\right|.
\end{equation}
While this metric is widely used, cumulative statistics can mask local miscalibration due to cancellation effects across confidence ranges.
\par Overall, uncertainty quality improves as $NLL$, $ECE_{\delta}$, and $ECE_{q}$ decrease. Calibration metrics in particular are not informative on datasets where accuracy is poor everywhere (\textit{e.g., } during a large domain shift). For this regime, we can visualize histograms of predictive entropy to compare the magnitude and distribution of uncertainty across methods.
\section{Experimental Setup}
\label{sec:implementation}
\begin{figure*}[t]
\centering
\includegraphics[width=1.0\textwidth]{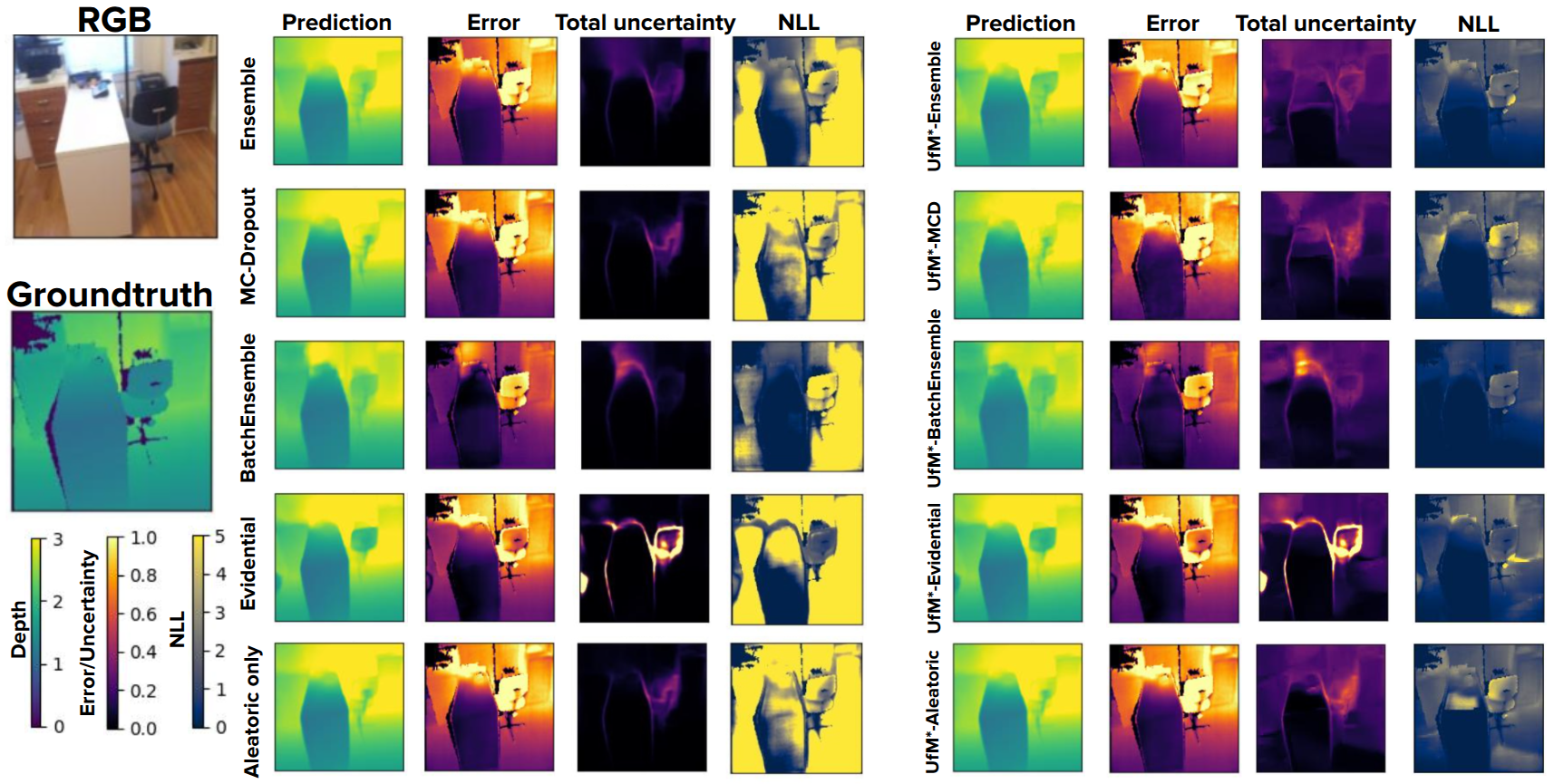}
\caption{Qualitative results on baselines and UfM$^*$ applied to these methods. While the baselines all capture some uncertainty in the image, they all are overconfident, resulting in a high NLL (lower is better). Meanwhile, incorporating multiview disagreement via UfM$^*$ across all methods results in uncertainty that better captures the error.}
\label{fig:qual_all_methods}
\end{figure*}
\par \textit{DNN architecture and training:} We evaluate performance using three different DNN architectures: ResNet50-Monodepth2~\cite{godard2019digging}, FastDepth~\cite{wofk2019fastdepth}, and Depth Anything V2-small~\cite{yang2024depth}. We train the following models for ResNet50-Monodepth2:
\begin{itemize}
    \item \textit{Ensemble:} an ensemble of size $N=10$ with NLL loss~\cite{kendall2017uncertainties} where each model predicts depth and aleatoric variance. 
    \item \textit{MC-Dropout:} a model with NLL loss~\cite{kendall2017uncertainties} that predicts depth and aleatoric variance. At test time, we apply MC-Dropout only to the decoder as in prior work~\cite{poggi2020uncertainty} and show results for 10 samples per inference for the baseline.
    \item \textit{BatchEnsemble:} a BatchEnsemble of size $N=10$ with NLL loss~\cite{kendall2017uncertainties} where each model predicts depth and aleatoric variance. 
    \item \textit{Evidential:} a model with evidential loss~\cite{amini2020deep} that predicts depth, aleatoric variance, and epistemic variance.
    \item \textit{Aleatoric:} a model with NLL loss~\cite{kendall2017uncertainties} that predicts depth and aleatoric variance. 
\end{itemize}
For all ResNet50-Monodepth2 and FastDepth models, we train for 50 epochs on the NYUDepthV2 dataset~\cite{silberman2012indoor} with resized 224$\times$224 inputs using Adam optimizer with 0.9 momentum, 0.0001 weight decay, and dropout ($p=0.2$) layers. Since we use the pretrained, knowledge distilled Depth Anything V2-small model~\cite{yang2024depth}, we test methods that only require predicted depth without any additional training on 518$\times$518 resized inputs. For brevity, we refer to it as Depth Anything V2 for the remainder of the article. We report $\delta_{1}$ accuracy as the proportion of pixels with predicted depth within 25\% of the ground-truth depth. 
\par \textit{Hyperparameters for UfM$^*$:} Unless otherwise specified, we use the following key hyperparameters in the experiments. Regarding segmentation, we use SPGF default hyperparameters~\cite{li2022memory} for depth-adaptive segmentation, and prune Gaussians that are too small (contain fewer than 2000 pixels or less than 32 rows). Regarding correspondences, we use overlap threshold to be $\eta = 0.5$ and occlusion filtering at $\eta_{occl} = 0.7$. To make Gaussian mixture regression more efficient, for each point, we search for nearby Gaussians in an R-tree~\cite{guttman1984r} up to $0.1$ meters away, and expand to $0.5$ meters if no Gaussians are found. We set the prior weight $\pi_{0} = 1$ with prior uncertainty $\mu_{0} = 0$. Unless otherwise specified, exponential moving average (EMA) smoothing with $\alpha = 0.5$ is applied to stabilize uncertainty estimates over time.
\par \textit{Datasets:} We test on 100 randomly sampled ScanNet~\cite{dai2017scannet} sequences to test uncertainty quality on an out-of-distribution dataset. We additionally test on forest and city environments in the TartanAir~\cite{wang2020tartanair} dataset and driving sequences in KITTI-360~\cite{liao2022kitti} dataset to evaluate performance on datasets with extreme distribution shift and scale difference. Finally, the system demonstration on a custom miniature car is run using data collected by the car from its camera in a motion capture room with objects placed on the ground.
\par \textit{Hardware:} For uncertainty quality and memory metrics, we run on an NVIDIA GeForce RTX 2080 Ti and a single core Intel(R) Xeon(R) Gold 6252 CPU @ 2.10GHz for reproducibility. For latency, power, and energy metrics, we run experiments on and measure energy using NVIDIA Jetson AGX Orin Developer Kit in 30 W mode on both the embedded GPU and four cores of the CPU. 

\section{Experimental Results}
\label{sec:experimental_results}
\newcolumntype{C}[1]{>{\centering\arraybackslash}p{#1}}
\newcolumntype{L}[1]{>{\raggedright\arraybackslash}p{#1}}
\newcolumntype{Y}{>{\raggedright\arraybackslash}X} 
\newcolumntype{A}{>{\centering\arraybackslash}m{1.6cm}} 
\newcolumntype{M}{>{\raggedright\arraybackslash}m{2.35cm}} 
\par In this section, we present experimental results evaluating UfM$^*$ on a diverse set of datasets to assess both uncertainty quality and efficiency. Specifically, we evaluate improvements in uncertainty quality in Sec.~\ref{sec:exp_results_quality}, memory and energy consumption in Sec.~\ref{sec:exp_results_eff}, the use of UfM$^*$ for post-hoc uncertainty estimation for a pretrained DNN in Sec.~\ref{sec:exp_results_post_hoc}, limitations under extreme distribution shift in Sec.~\ref{sec:exp_results_dist_shift}, and ablations analyzing the effects of pose noise, image overlap, and multiple inferences in Sec.~\ref{sec:exp_results_pose_noise_overlap_robustness}. Finally, in Sec.~\ref{sec:exp_results_system_demo}, we demonstrate UfM$^*$ deployed fully onboard a resource-constrained miniature autonomous car platform using real-world data.
\par Throughout this section, we compare against and use as the base network for multiview methods the following commonly used uncertainty estimation baselines: ensembles, MC-Dropout, BatchEnsemble, evidential, aleatoric (specific implementations described in Sec.~\ref{sec:implementation}). We additionally compare against UfM~\cite{sudhakar2022uncertainty}, a pointwise multiview disagreement uncertainty estimation algorithm that can also be applied to the base networks above.  
While not exhaustive, this set of baselines spans representative approaches for ensemble-based, sampling-based, single model, and multiview uncertainty methods.

\subsection{Uncertainty quality} \label{sec:exp_results_quality}

\begin{figure*}[t]
\centering
\includegraphics[width=0.9\textwidth]{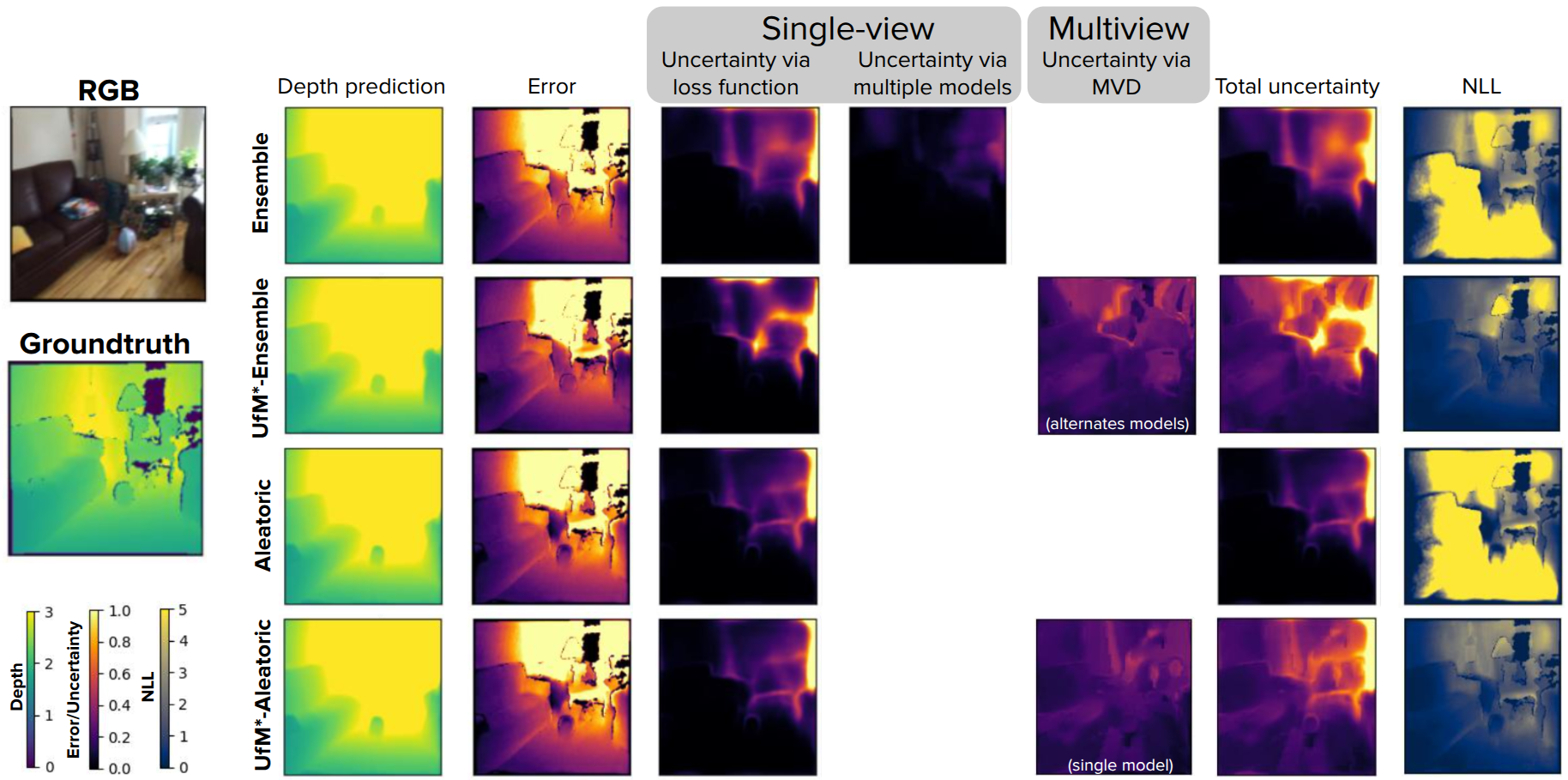} %
\caption{Visualization of uncertainty induced by loss function, multiple models, and multiview disagreement (MVD). We see that MVD is able to capture uncertainty in areas where other uncertainty measurement methods are confidently wrong.
}
\label{fig:qual_breakdown}
\end{figure*}

\begin{figure}[t]
\centering
\includegraphics[width=1.0\columnwidth]{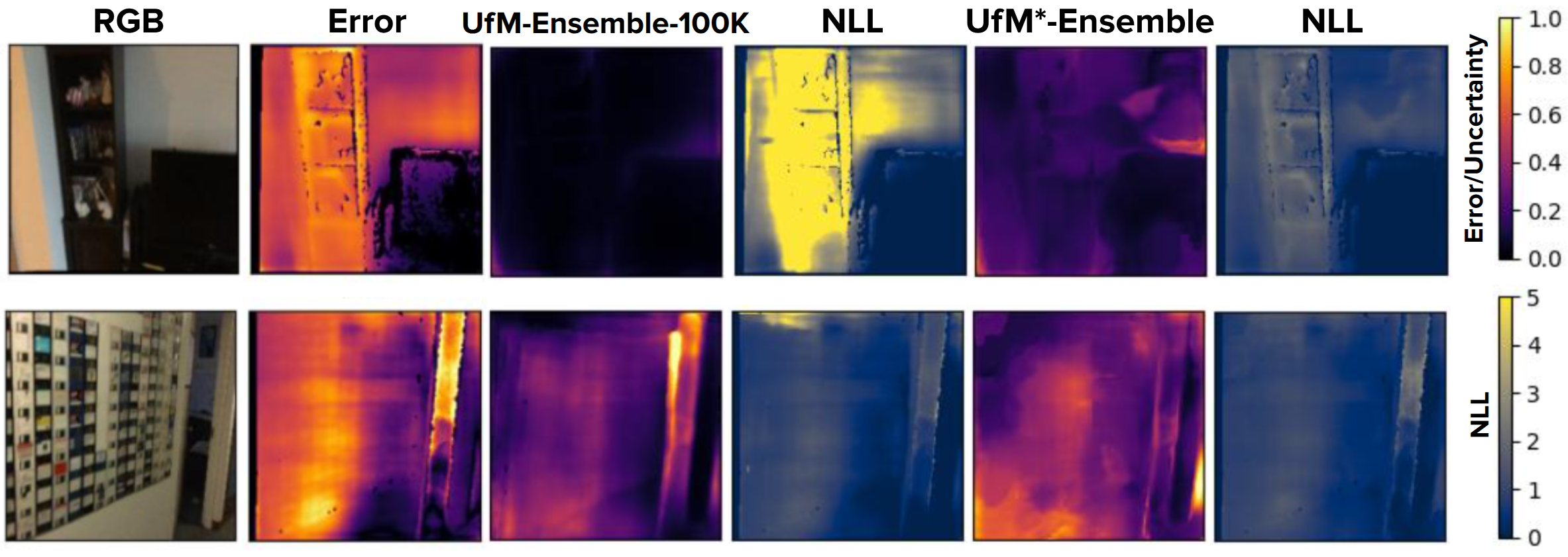} 
\caption{UfM$^*$-Ensemble vs. UfM-Ensemble-100K. 
}
\label{fig:qual_ufmstar_vs_ufm}
\end{figure}
\par \textit{Qualitative results:} Fig.~\ref{fig:qual_all_methods} shows qualitative examples of error, total uncertainty, and NLL for the baselines and their UfM$^*$ variants. As we can see, across all methods, incorporating multiview disagreement through UfM$^*$ consistently improves uncertainty quality by reducing overconfidence.
Fig.~\ref{fig:qual_breakdown} illustrates the decomposition of total uncertainty into aleatoric, epistemic, and multiview components for ensemble and aleatoric baselines and their UfM$^*$ variants. While aleatoric uncertainty can identify regions of high error, it is overconfident in some areas of high error as seen by the high NLL. UfM$^*$ captures geometric inconsistency across views, increasing uncertainty in these regions and improving uncertainty quality (lower NLL) compared to both aleatoric and ensemble baselines.
Fig.~\ref{fig:qual_ufmstar_vs_ufm} compares UfM$^*$-Ensemble with its pointwise counterpart, UfM-Ensemble-100K, where 100K denotes the pruning threshold used to make the point cloud computationally tractable. Multiview disagreement in UfM$^*$ reduces overconfidence (\textit{e.g.}, the bookcase in the top row), though it slightly reduces sensitivity to fine-grained details compared to pointwise multiview disagreement (\textit{e.g.,} doorway in bottom row).
\begin{table*}[t]
\vspace{0.2cm}
\centering
\caption{Uncertainty quality and efficiency averaged over 100 ScanNet sequences with 95\% confidence intervals (CI) of the mean (standard error) using ResNet50-Monodepth2 (CIs smaller than $0.01$ are omitted). Accuracy 95\% CI within $\pm 0.03$. Note, storing previous uncertainty map for smoothing requires additional 0.2 MB of memory for UfM$^*$ methods.}
\label{tab:benchmarks_uq_and_eff}
\setlength{\tabcolsep}{3pt}
\renewcommand{\arraystretch}{1.15}

{\footnotesize
\begin{tabularx}{\textwidth}{
    Y
    C{1.5cm}
    C{1.5cm}
    C{1.5cm}
    C{1.5cm}
    C{1.7cm}
    C{1.7cm}
}
\toprule
& Accuracy & \multicolumn{3}{c}{Uncertainty quality} & \multicolumn{2}{c}{Uncertainty efficiency} \\
\cmidrule(lr){2-2}\cmidrule(lr){3-5}\cmidrule(lr){6-7}
Method & $\delta_{1}$ ($\uparrow$)& NLL ($\downarrow$) & ECE$_{\delta}$ ($\downarrow$)& ECE$_{q}$ ($\downarrow$)& Energy [mJ]& Memory [MB]\\
\midrule

Ensemble~\cite{lakshminarayanan2017simple} & $0.70$ & $1.33 \pm 0.17$ & $0.17 \pm 0.03$ & $0.18 \pm 0.02$ & $1857 \pm 2$ & $\mathbf{1245}$ \\
UfM-Ensemble~\cite{sudhakar2022uncertainty} & $0.68$ & $1.34 \pm 0.22$ & $0.16 \pm 0.02$ & $0.16 \pm 0.02$ & $257 \pm 18$ & $1319 \pm 7$ \\
UfM-Ensemble-100K~\cite{sudhakar2022uncertainty} & $0.68$ & $1.51 \pm 0.23$ & $0.17 \pm 0.02$ & $0.17 \pm 0.02$ & $112 \pm 2$ & $1249 $ \\
\textbf{UfM$^*$-Ensemble} & $0.68$ & $\mathbf{0.96} \pm 0.19$ & $\mathbf{0.12} \pm 0.02$ & $\mathbf{0.13} \pm 0.02$ & $\mathbf{66} \pm 2$ & $\mathbf{1245} $ \\
\addlinespace

MC-Dropout~\cite{kendall2017uncertainties} & $0.68$ & $3.87 \pm 0.47$ & $0.24 \pm 0.03$ & $0.19 \pm 0.02$ & $1854 \pm 2$ & $\mathbf{0.01}$ \\
UfM-MCD~\cite{sudhakar2022uncertainty} & $0.68$ & $1.95 \pm 0.28$ & $0.18 \pm 0.02$ & $0.16 \pm 0.02$ & $242 \pm 16$ & $76 \pm 7$ \\
UfM-MCD-100K~\cite{sudhakar2022uncertainty} & $0.68$ & $2.33 \pm 0.31$ & $0.20 \pm 0.02$ & $0.17 \pm 0.02$ & $111 \pm 2$ & $4$ \\
\textbf{UfM$^*$-MCD} & $0.68$ & $\mathbf{1.12} \pm 0.18$ & $\mathbf{0.13} \pm 0.02$ & $\mathbf{0.13} \pm 0.02$ & $\mathbf{61} \pm 2$ & $0.03 $ \\
\addlinespace

BatchEnsemble~\cite{wen2019batchensemble} & $0.68$ & $2.15 \pm 0.26$ & $0.22 \pm 0.02$ & $0.17 \pm 0.02$ & $1580 \pm 3$ & $\mathbf{2}$ \\
UfM-BatchEnsemble~\cite{sudhakar2022uncertainty} & $0.67$ & $1.48 \pm 0.24$ & $0.18 \pm 0.02$ & $0.15 \pm 0.02$ & $302 \pm 16$ & $80 \pm 7$ \\
UfM-BatchEnsemble-100K~\cite{sudhakar2022uncertainty} & $0.67$ & $1.74 \pm 0.26$ & $0.20 \pm 0.02$ & $0.16 \pm 0.02$ & $162 \pm 2$ & $6 $ \\
\textbf{UfM$^*$-BatchEnsemble} & $0.67$ & $\mathbf{1.01} \pm 0.18$ & $\mathbf{0.11} \pm 0.02$ & $\mathbf{0.12} \pm 0.02$ & $\mathbf{123} \pm 1$ & $\mathbf{2} $ \\
\addlinespace

Evidential~\cite{amini2020deep} & $0.67$ & $3.51 \pm 0.38$ & $0.22 \pm 0.02$ & $0.16 \pm 0.02$ & $\mathbf{30} \pm 1$ & $\mathbf{0.03}$ \\
UfM-Evidential~\cite{sudhakar2022uncertainty} & $0.67$ & $1.85 \pm 0.22$ & $0.19 \pm 0.02$ & $0.14 \pm 0.02$ & $245 \pm 13$ & $75 \pm 7$ \\
UfM-Evidential-100K~\cite{sudhakar2022uncertainty} & $0.67$ & $2.17 \pm 0.25$ & $0.20 \pm 0.02$ & $0.15 \pm 0.02$ & $119 \pm 2$ & $4 $ \\
\textbf{UfM$^*$-Evidential} & $0.67$ & $\mathbf{1.12} \pm 0.15$ & $\mathbf{0.16} \pm 0.01$ & $\mathbf{0.13} \pm 0.02$ & $74 \pm 2$ & $0.05$ \\
\addlinespace

Aleatoric~\cite{kendall2017uncertainties} & $0.68$ & $4.10 \pm 0.50$ & $0.24 \pm 0.03$ & $0.19 \pm 0.02$ & $\mathbf{10} \pm 1$ & $\mathbf{0.01}$ \\
UfM-Aleatoric~\cite{sudhakar2022uncertainty} & $0.68$ & $1.99 \pm 0.28$ & $0.18 \pm 0.02$ & $0.16 \pm 0.02$ & $234 \pm 13$ & $76 \pm 7$ \\
UfM-Aleatoric-100K~\cite{sudhakar2022uncertainty} & $0.68$ & $2.38 \pm 0.31$ & $0.20 \pm 0.02$ & $0.17 \pm 0.02$ & $109 \pm 2$ & $4$ \\
\textbf{UfM$^*$-Aleatoric} & $0.68$ & $\mathbf{1.11} \pm 0.18$ & $\mathbf{0.13} \pm 0.02$ & $\mathbf{0.13} \pm 0.02$ & $61 \pm 2$ & $0.04$ \\
\bottomrule
\end{tabularx}
}
\end{table*}
\begin{figure}[t!] \centering
         \centering
         \includegraphics[width=0.49\columnwidth]{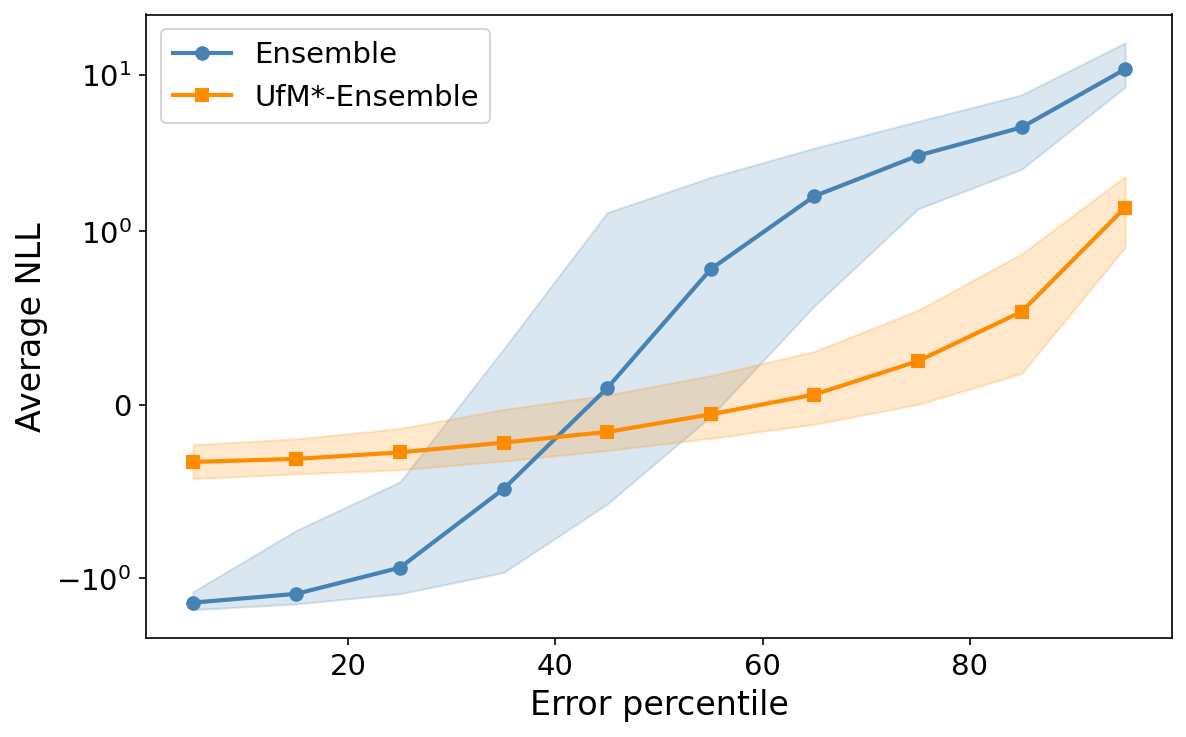}
         \includegraphics[width=0.49\columnwidth]{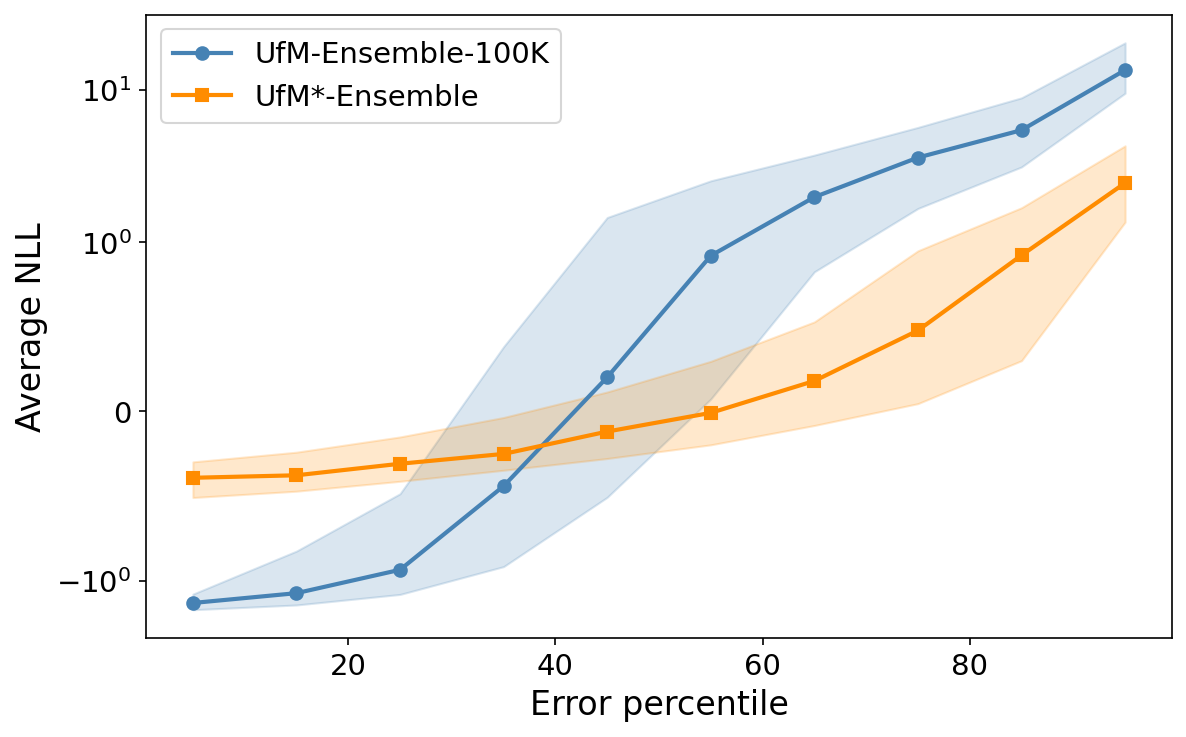}
     \caption{Left: comparison against ensemble confident pixels, right: comparison against UfM-Ensemble confident pixels. We see UfM$^*$ is able to improve uncertainty quality on overconfident failures while introducing some underconfidence for accurate predictions.}
     \label{fig:binned_error_vs_nll}
\end{figure}

\par \textit{Results across 100 ScanNet sequences:} To evaluate whether these qualitative improvements generalize across the dataset, we report average results over 100 ScanNet sequences in Table~\ref{tab:benchmarks_uq_and_eff} for all baselines and their UfM$^*$ variants for ResNet50-Monodepth2-based architectures. Across all baselines, incorporating UfM$^*$ improves uncertainty quality relative to the corresponding baseline method. UfM$^*$-Ensemble achieves the best overall uncertainty quality, surpassing ensemble uncertainty quality while requiring only 3\% of the energy.
\par To better understand the source of the performance gains and limitations of UfM$^*$ relative to ensembles, we analyze pixels where the ensemble uncertainty is low (confident) and group them by error percentile across all sequences (Fig.~\ref{fig:binned_error_vs_nll}). UfM$^*$ reduces NLL primarily by correcting overconfident predictions, though this comes at the cost of introducing some underconfidence. A similar trend is observed when comparing UfM$^*$ to a UfM variant: UfM$^*$ reduces overconfidence but introduces mild underconfidence in certain regions.
\begin{figure}[t]
\centering
\includegraphics[width=0.3\columnwidth]{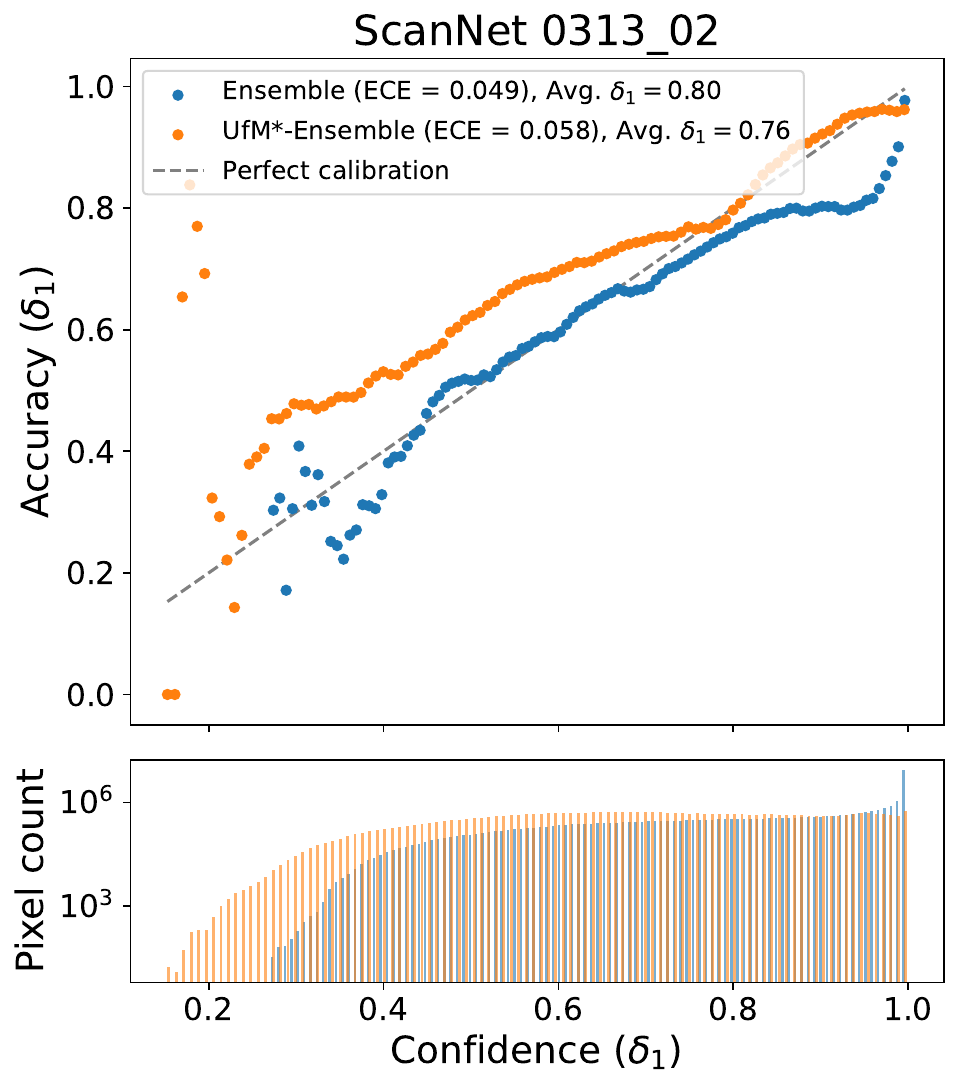} 
\includegraphics[width=0.3\columnwidth]{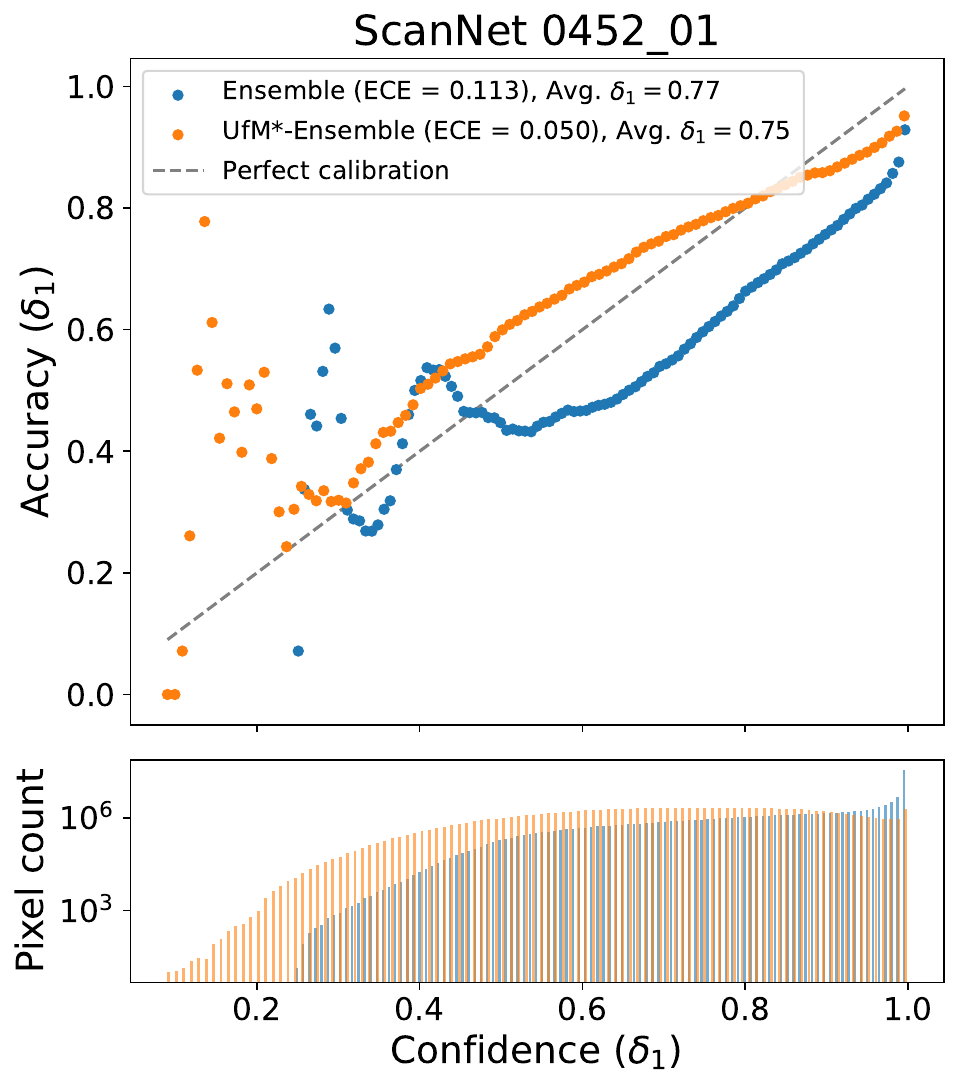} 
\includegraphics[width=0.3\columnwidth]{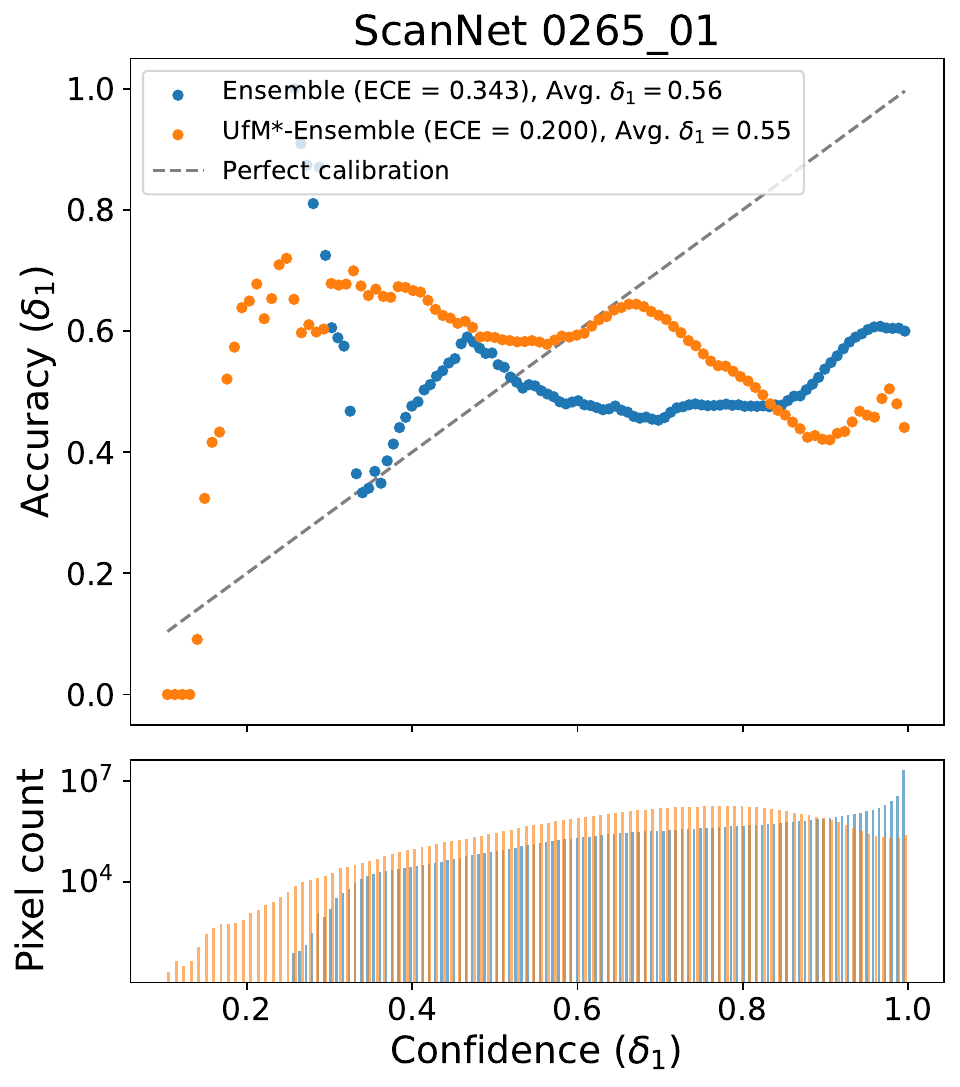}
\includegraphics[width=0.3\columnwidth]{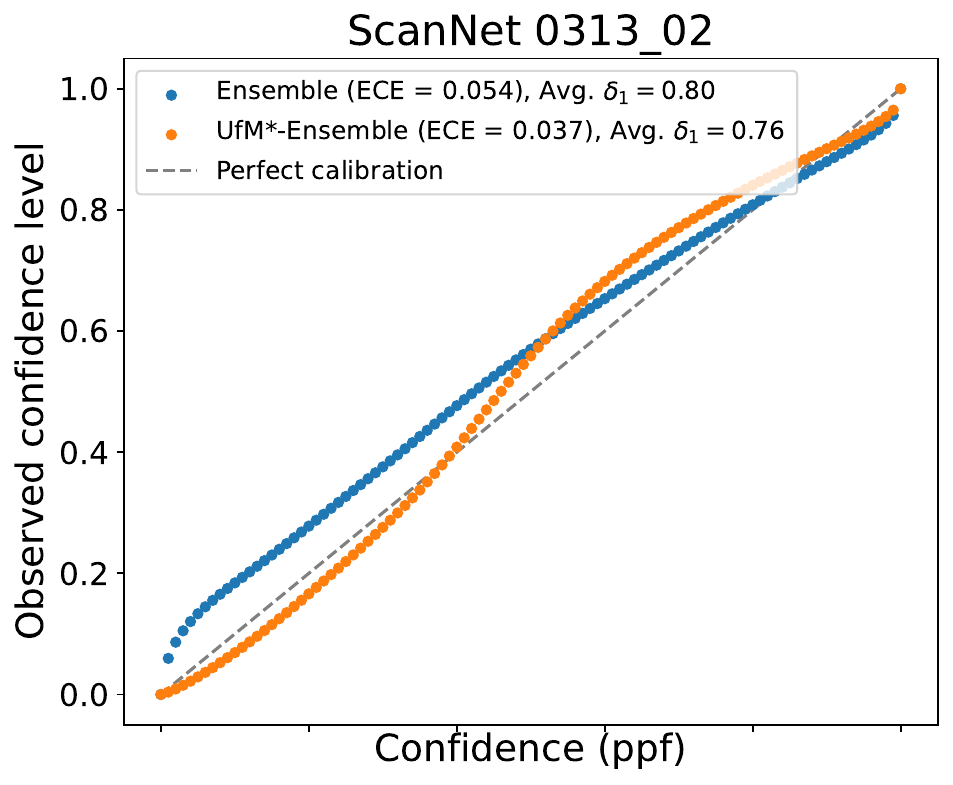} 
\includegraphics[width=0.3\columnwidth]{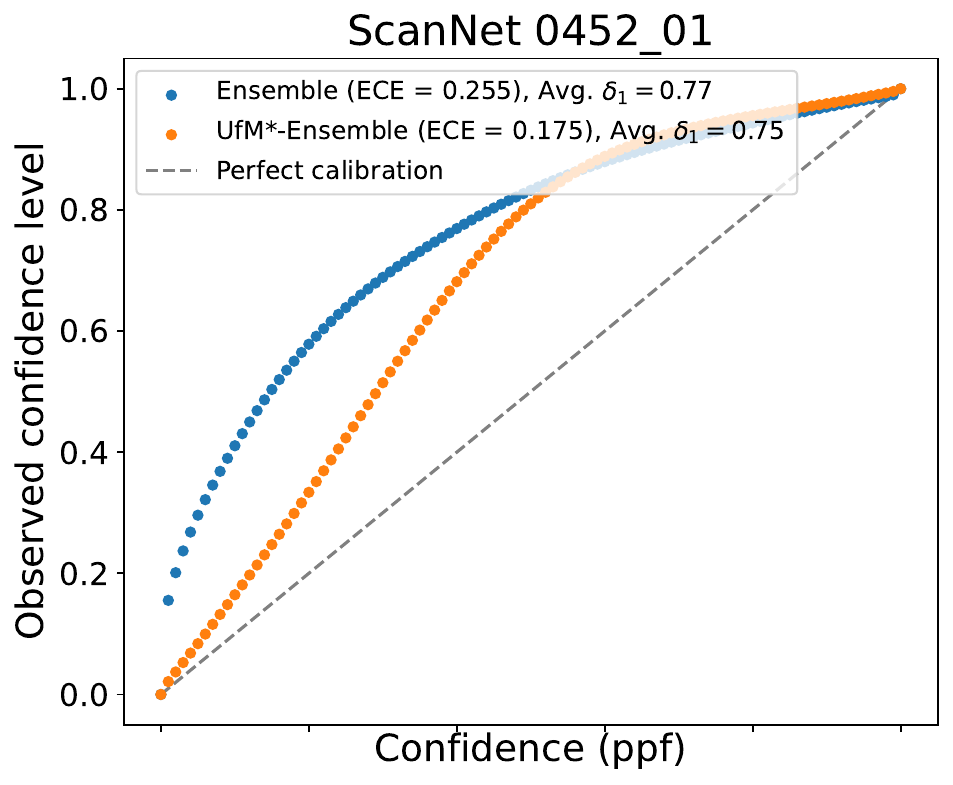} 
\includegraphics[width=0.3\columnwidth]{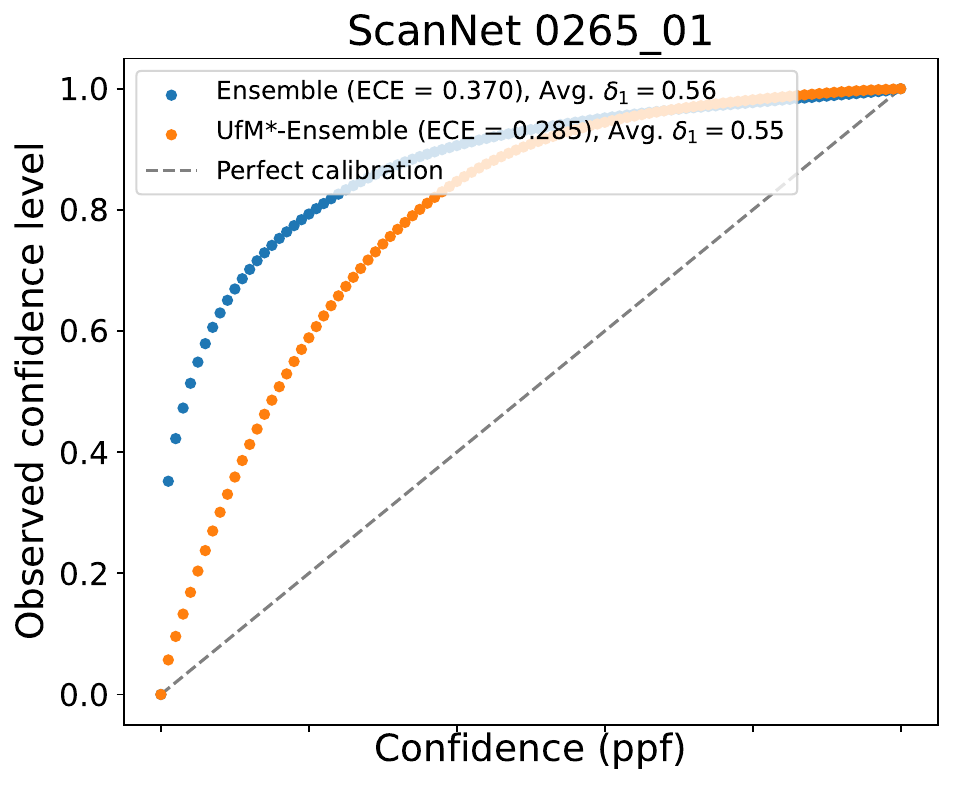}
\caption{Calibration curves for three different ScanNet sequences (top row: $\delta$ calibration, bottom row: quantile calibration). First column: both ensemble and UfM$^*$-Ensemble well-calibrated. Second column: ensemble poorly calibrated and UfM$^*$-Ensemble well-calibrated. Third column: both ensemble and UfM$^*$-Ensemble poorly calibrated.}
\label{fig:calibration_curves}
\end{figure} 
\subsection{Compute- and memory efficiency of computing uncertainty} \label{sec:exp_results_eff}
\begin{table*}[t]
\vspace{0.2cm}
\centering
\caption{Breakdown of latency, power, energy, and memory for uncertainty estimation on an Nvidia Orin AGX platform, averaged across 100 ScanNet sequences. MVD memory measures the memory of the 3D representation (\textit{e.g.,} GMM, point cloud). Note, storing previous uncertainty map for smoothing requires additional 0.2 MB of memory for UfM$^*$ methods.}
\label{tab:efficiency_in_detail}

\setlength{\tabcolsep}{2pt}
\renewcommand{\arraystretch}{1.12}

\newcommand{\ArchCellResNet}{%
\begin{tabular}{@{}c@{}}
ResNet50\\
Monodepth2\\[-1pt]
{\scriptsize Lat.: 27 ms}\\[-1pt]
{\scriptsize Mem.: 138.3 MB}
\end{tabular}}

\newcommand{\ArchCellFD}{%
\begin{tabular}{@{}c@{}}
FastDepth\\[-1pt]
{\scriptsize Lat.: 12 ms}\\[-1pt]
{\scriptsize Mem.: 15.8 MB}
\end{tabular}}

\newcommand{\MethodOneLine}[1]{{\fontsize{7}{8}\selectfont\mbox{#1}}}

{\footnotesize
\begin{tabularx}{\textwidth}{
    C{1.75cm}      
    M              
    C{0.85cm}      
    C{0.85cm}      
    C{0.85cm}      
    C{0.85cm}      
    C{0.85cm}      
    C{0.85cm}      
    C{1.75cm}      
    C{2cm}      
    C{1.5cm}      
    C{1.75cm}      
}
\toprule
& & \multicolumn{7}{c}{Latency / Energy} & \multicolumn{3}{c}{Memory} \\
\cmidrule(lr){3-9}\cmidrule(lr){10-12}
& & \multicolumn{3}{c}{GPU} & \multicolumn{3}{c}{CPU} & & & & \\
\cmidrule(lr){3-5}\cmidrule(lr){6-8}
Arch & Method
& Latency [ms]
& Power [W]
& Energy [mJ]
& Latency [ms]
& Power [W]
& Energy [mJ]
& Energy overhead [\%]
& DNN unc. mem. [MB]
& MVD mem. [MB]
& Mem. overhead [\%] \\
\midrule

\multirow{8}{=}{\ArchCellResNet}
& \MethodOneLine{Ensemble~\cite{lakshminarayanan2017simple}}        & $252$ & $7.4$ & $1857$ & $0$ & $0$ & $0$ & $90$ & $1245$ & -- & $90.0$ \\
& \MethodOneLine{UfM-Ensemble~\cite{sudhakar2022uncertainty}}       & $2$ & $4.4$ & $9$ & $128$ & $1.9$ & $248$ & $67$ & $1245$ & $74.6$ & $90.5$ \\
& \MethodOneLine{UfM-Ensemble-100K~\cite{sudhakar2022uncertainty}}  & $2$ & $4.7$ & $10$ & $59$ & $1.7$ & $102$ & $47$ & $1245$ & $4.0$ & $90.0$ \\
& \MethodOneLine{\textbf{UfM$^*$-Ensemble}}                         & $2$ & $5.2$ & $13$ & $24$ & $2.1$ & $53$ & $32$ & $1245$ & $0.03$ & $90.0$ \\
\addlinespace
& \MethodOneLine{Aleatoric~\cite{kendall2017uncertainties}}         & $2$ & $6.6$ & $10$ & $0$ & $0$ & $0$ & $6$ & $0.01$ & -- & $0.01$ \\
& \MethodOneLine{UfM-Aleatoric}                                     & $2$ & $4.5$ & $7$ & $116$ & $1.9$ & $226$ & $65$ & $0.01$ & $75.9$ & $33.9$ \\
& \MethodOneLine{UfM-Aleatoric-100K}                                & $2$ & $4.7$ & $8$ & $58$ & $1.7$ & $101$ & $46$ & $0.01$ & $4.0$ & $2.8$ \\
& \MethodOneLine{\textbf{UfM$^*$-Aleatoric}}                        & $2$ & $5.3$ & $11$ & $23$ & $2.1$ & $50$ & $30$ & $0.01$ & $0.03$ & $0.03$ \\
\midrule

\multirow{8}{=}{\ArchCellFD}
& \MethodOneLine{Ensemble~\cite{lakshminarayanan2017simple}}        & $112$ & $5.4$ & $608$ & $0$ & $0$ & $0$ & $90$ & $142.6$ & -- & $90.0$ \\
& \MethodOneLine{UfM-Ensemble~\cite{sudhakar2022uncertainty}}       & $<1$ & $3.9$ & $1$ & $113$ & $1.9$ & $217$ & $81$ & $142.6$ & $75.3$ & $93.1$ \\
& \MethodOneLine{UfM-Ensemble-100K~\cite{sudhakar2022uncertainty}}  & $<1$ & $3.9$ & $1$ & $61$ & $1.6$ & $96$ & $66$ & $142.6$ & $4.0$ & $90.3$ \\
& \MethodOneLine{\textbf{UfM$^*$-Ensemble}}                         & $1$ & $4.1$ & $3$ & $24$ & $2.3$ & $56$ & $53$ & $142.6$ & $0.03$ & $90.0$ \\
\addlinespace
& \MethodOneLine{Aleatoric~\cite{kendall2017uncertainties}}         & $<1$ & $4.6$ & $2$ & $0$ & $0$ & $0$ & $3$ & $< 0.01$ & -- & $< 0.01$ \\
& \MethodOneLine{UfM-Aleatoric}                                     & $<1$ & $3.9$ & $1$ & $114$ & $1.9$ & $220$ & $81$ & $< 0.01$ & $77.4$ & $80.4$ \\
& \MethodOneLine{UfM-Aleatoric-100K}                                & $<1$ & $3.9$ & $1$ & $60$ & $1.6$ & $95$ & $66$ & $< 0.01$ & $4.0$ & $20.2$ \\
& \MethodOneLine{\textbf{UfM$^*$-Aleatoric}}                        & $1$ & $4.1$ & $2$ & $23$ & $2.3$ & $53$ & $52$ & $< 0.01$ & $0.03$ & $0.2$ \\
\bottomrule
\end{tabularx}
}
\end{table*}
\begin{figure}[t!] \centering
         \centering
         \includegraphics[width=0.9\columnwidth]{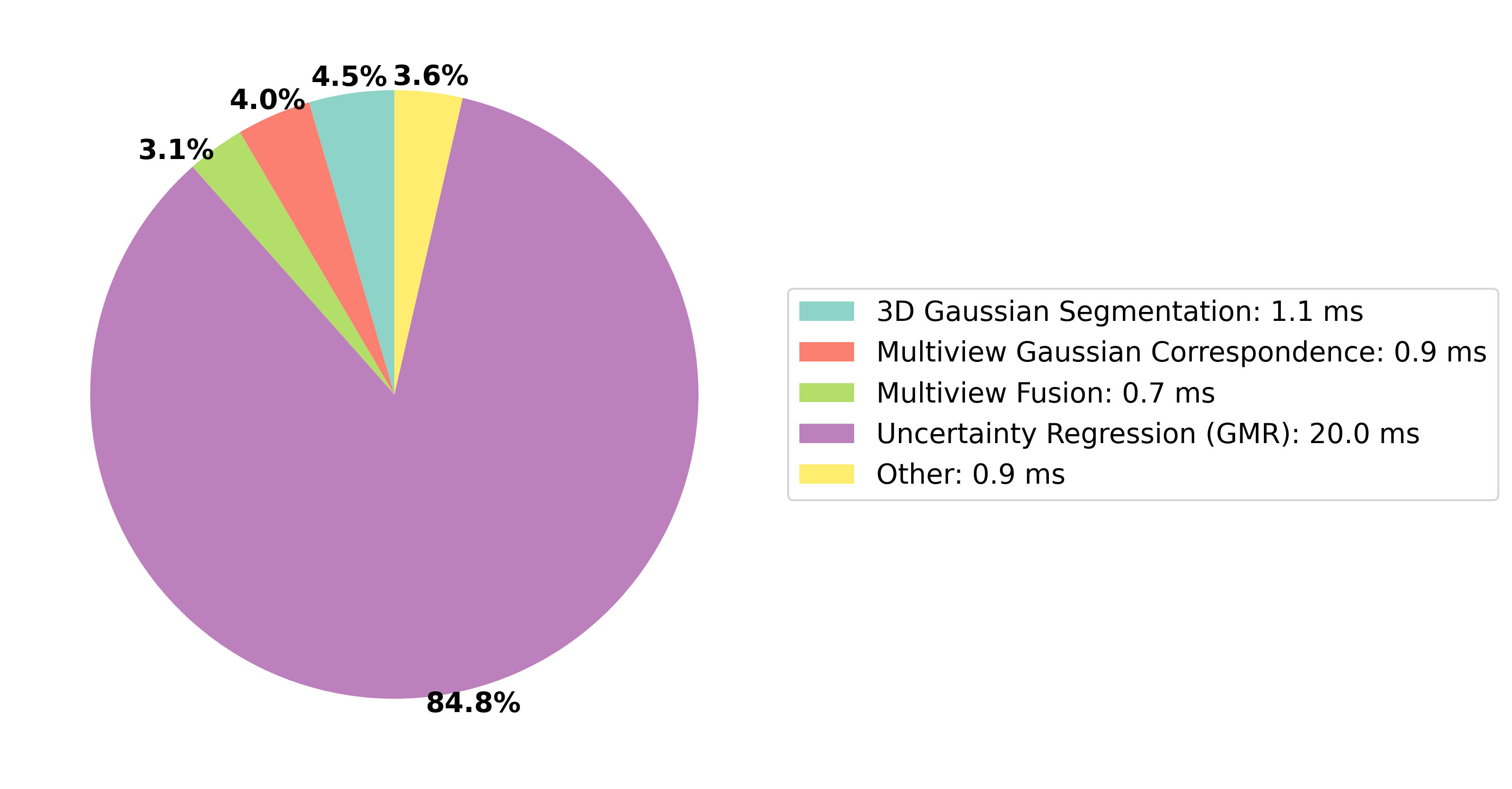}
     \caption{Average latency percentages for UfM$^*$-Ensemble method on 100 ScanNet sequences (24 ms per image total); uncertainty regression using GMR dominates overhead.}
     \label{fig:overhead_breakdown}
\end{figure}
\par As shown in Table~\ref{tab:benchmarks_uq_and_eff}, while UfM$^*$-Ensemble achieves the best uncertainty quality and is energy-efficient, it is not memory-efficient because it still requires maintaining multiple ensemble members and alternating between them during inference. For resource-constrained systems where both compute and memory are limited, alternative variants such as UfM$^*$-BatchEnsemble (memory-efficient ensemble), UfM$^*$-MCD (sampling-based), and single model approaches such as UfM$^*$-Evidential and UfM$^*$-Aleatoric provide more favorable efficiency trade-offs. For instance, UfM$^*$-Aleatoric reduces $ECE_{\delta}$ and $ECE_{q}$ by 24-28\% compared to an ensemble while consuming only $3$\% of the energy and $0.02$\% of the memory.

\par To better understand these trade-offs, Table~\ref{tab:efficiency_in_detail} provides a detailed breakdown of latency, power, energy, and memory usage for ensemble and aleatoric baselines and their UfM and UfM$^*$ variants using both a ResNet50-Monodepth2~\cite{godard2019digging} architecture and the lightweight FastDepth~\cite{wofk2019fastdepth}. In all methods, the DNN components responsible for uncertainty estimation (e.g., aleatoric or evidential heads, additional ensemble members) run on the GPU, while the multiview component (UfM or UfM$^*$) runs on two CPU cores. Energy and memory overhead are defined as the fraction of energy or memory consumed by uncertainty estimation relative to the total cost of both depth prediction and uncertainty estimation. 
\par Clearly, using an ensemble is the most expensive. For an ensemble of size 10, uncertainty estimation accounts for approximately $90\%$ of the total latency, energy, and memory overhead. In contrast, the aleatoric baseline is the cheapest approach in terms of both energy and memory, though as shown in Table~\ref{tab:benchmarks_uq_and_eff} it provides lower uncertainty quality.
We next compare UfM$^*$ to the UfM variants. Without pruning, UfM-Ensemble and UfM-Aleatoric incur extremely high latency, energy, and memory costs, as the accumulated point cloud grows to millions of points in room-scale environments. Even when applying pruning to maintain a fixed point cloud size of $100$K points, UfM-Aleatoric-100K still incurs significant memory overhead. This is particularly noticeable for compact architectures such as FastDepth, where storing the point cloud alone requires $20\%$ of the total memory, which can be costly for off-chip memory access. In contrast, using a GMM dramatically reduces memory consumption while improving uncertainty quality. The GMM used by UfM$^*$ requires only tens of kilobytes on average and contributes less than $0.2\%$ of the total memory footprint for either architecture. 

\par Furthermore, by compressing multiview information into a compact set of Gaussians, UfM$^*$ enables real-time uncertainty estimation on CPU, running at over 40 FPS across the evaluated methods. Fig.~\ref{fig:overhead_breakdown} breaks down the computational cost of UfM$^*$ into its primary stages. The Gaussian mixture regression (GMR) stage dominates the latency because it operates per pixel and scales with the number of output pixels. However, this cost can be reduced depending on the application. For example, for an active learning algorithm that requires only an average uncertainty estimate per image~\cite{fu2025dectrain}, GMR can be evaluated on a subset of pixels to compute a sample average, significantly reducing computational cost.
\subsection{Post-hoc uncertainty using UfM$^*$} \label{sec:exp_results_post_hoc}
\begin{table}[t]
\vspace{0.1cm}
\centering
\caption{
Post-hoc uncertainty estimation for Depth Anything V2. 95\% CI for accuracy are $\pm 0.03$ and $<0.03$ for all ECE$_{\delta}$ and ECE$_{\textit{q}}$, and $<0.01$ for memory.}
\label{tab:depth_anything_v2}

\setlength{\tabcolsep}{3pt}
\renewcommand{\arraystretch}{1.1}

\begin{tabular}{l c c c c}
\toprule
Method
& $\delta_1\uparrow$ 
& ECE$_{\delta}$ $\downarrow$ 
& ECE$_{\textit{q}}$ $\downarrow$ 
& Mem. [MB] \\
\midrule
UfM-DepthOnly-300K                      & 0.76 & $0.23$ & $0.26$ & $12$ \\
UfM$^*$-DepthOnly-fixed-param           & 0.76 & $0.14$ & $0.16$ & $0.07$ \\
UfM$^*$-DepthOnly                       & 0.76 & $0.14$ & $0.17$ & $0.02$ \\
\bottomrule
\end{tabular}
\end{table}

\begin{figure}[t]
\centering
\includegraphics[width=0.49\columnwidth]{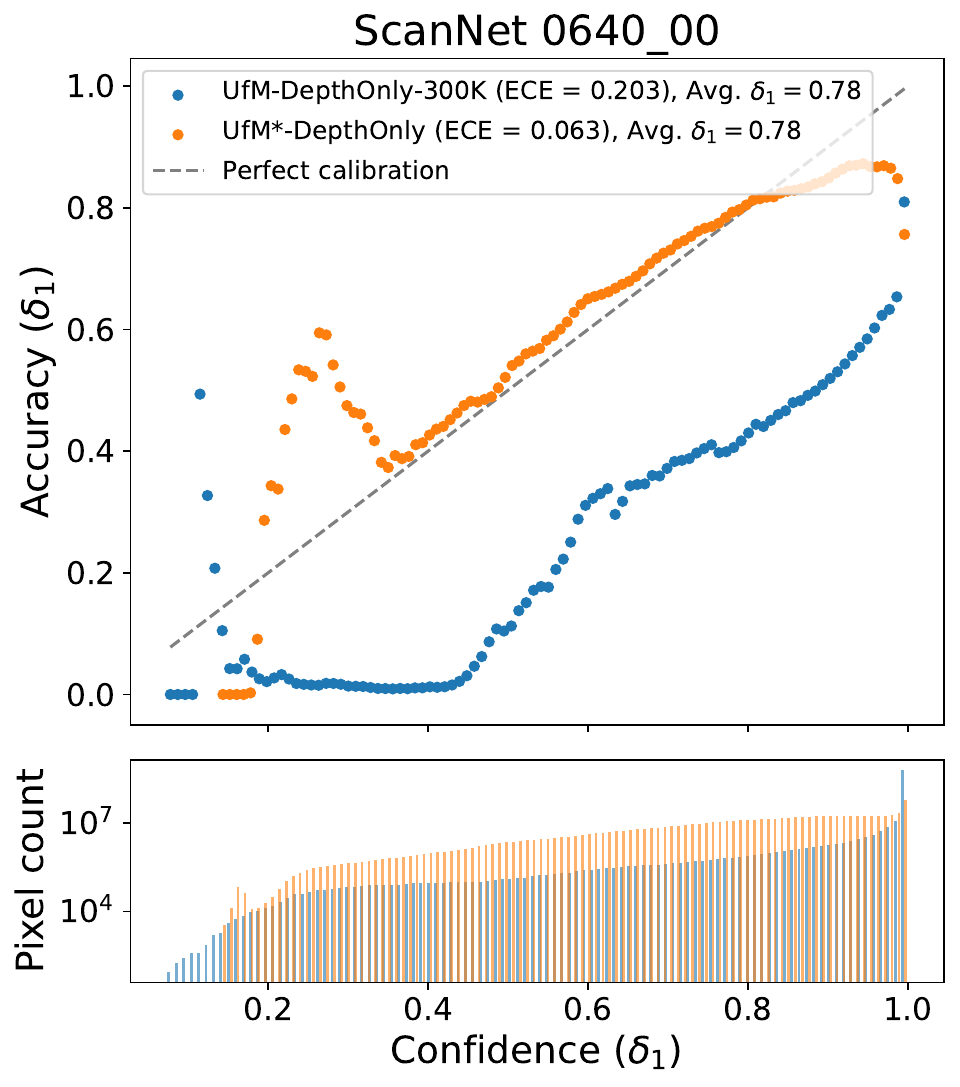} 
\includegraphics[width=0.49\columnwidth]{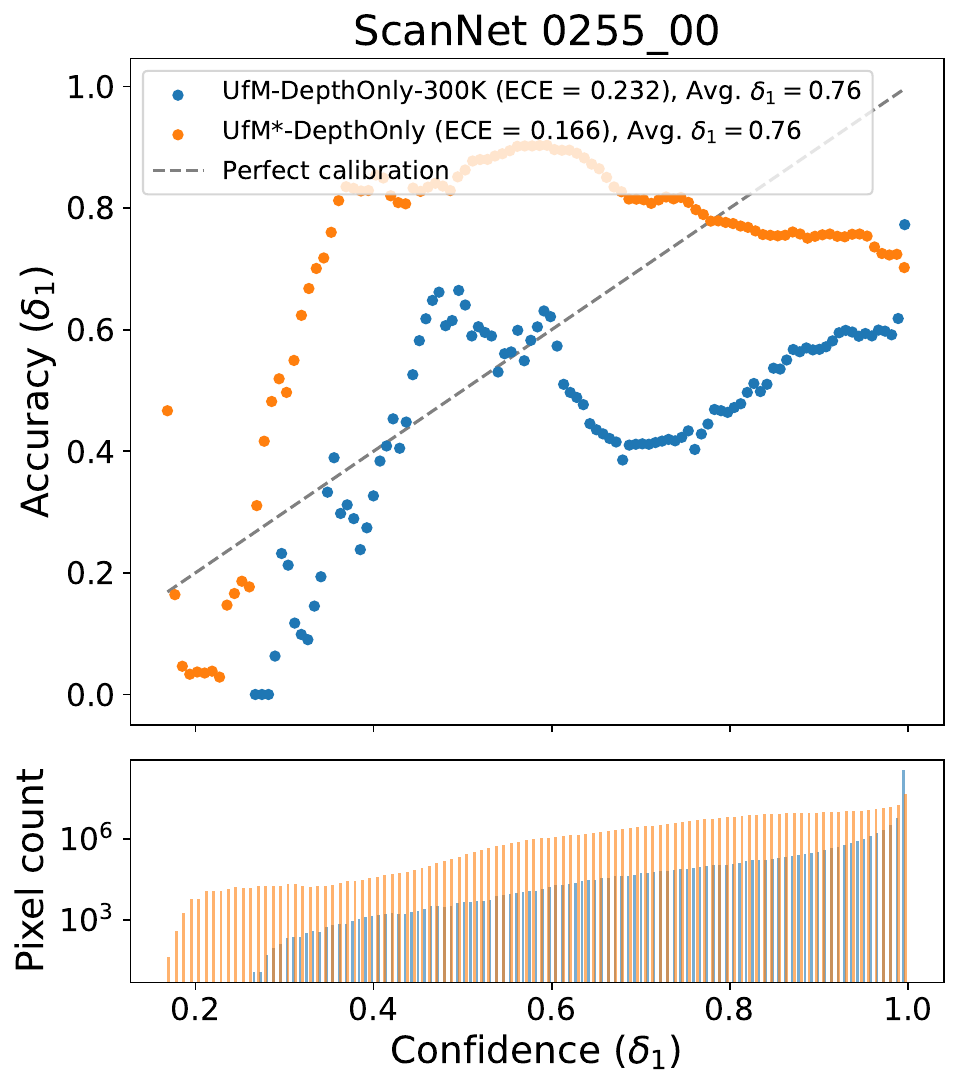} 
\caption{
Calibration curves for Depth Anything V2 model for UfM-DepthOnly-300K and UfM$^*$-DepthOnly. Left: we see UfM$^*$-DepthOnly gives a well-calibrated uncertainty signal with no modification to the model or additional training. Right: failure case where neither method is well-calibrated. 
}
\label{fig:calibration_curves_depth_anything_v2}
\end{figure}
\par Many uncertainty estimation methods require modifying the DNN architecture and retraining either a single model or an ensemble~\cite{lakshminarayanan2017simple, wen2019batchensemble, kendall2017uncertainties, amini2020deep, laurent2023packed, baumann2023probabilistic, charpentiernatural}. This can be challenging and computationally expensive for large models pretrained on massive datasets~\cite{yang2024depth, oquab2023Dinov2}. As a result, post-hoc approaches such as MC-Dropout remain popular despite requiring multiple inferences per image~\cite{gal2017deep}. Multiview disagreement methods (UfM$^*$, UfM) are also post-hoc and can be applied to any pretrained depth network using only its predicted depth outputs, but in contrast, only require a single inference per image.

\par To evaluate the ability to estimate uncertainty using predicted depth alone, we apply UfM and UfM$^*$ to a pretrained Depth Anything V2 model and evaluate on 100 ScanNet sequences, with results summarized in Table~\ref{tab:depth_anything_v2}. Because Depth Anything V2 operates effectively on larger inputs, we use an image resolution of $518\times518$ and increase the UfM pruning threshold to 300K points since the minimum pruning threshold is the number of pixels in the input. UfM$^*$-DepthOnly-fixed-param is first evaluated using the same hyperparameters used for $224\times224$ inputs, and then with scaled hyperparameters adapted for the larger resolution (UfM$^*$-DepthOnly). Both UfM and UfM$^*$ provide meaningful post-hoc uncertainty estimates using only predicted depth. However, UfM$^*$ achieves substantially higher uncertainty quality than UfM while requiring only a fraction of the memory footprint, which can be further adjusted through the scaled hyperparameters.

\par Representative calibration curves are shown in Fig.~\ref{fig:calibration_curves_depth_anything_v2}. In many cases, UfM$^*$-DepthOnly produces well-calibrated uncertainty estimates and corrects overconfident predictions produced by UfM-DepthOnly. However, uncertainty quality is not guaranteed, as illustrated by the example on the right where the calibration is poor for both methods.
\subsection{Out-of-distribution performance} \label{sec:exp_results_dist_shift}
\begin{figure}[t]
\centering
\includegraphics[width=0.7\columnwidth]{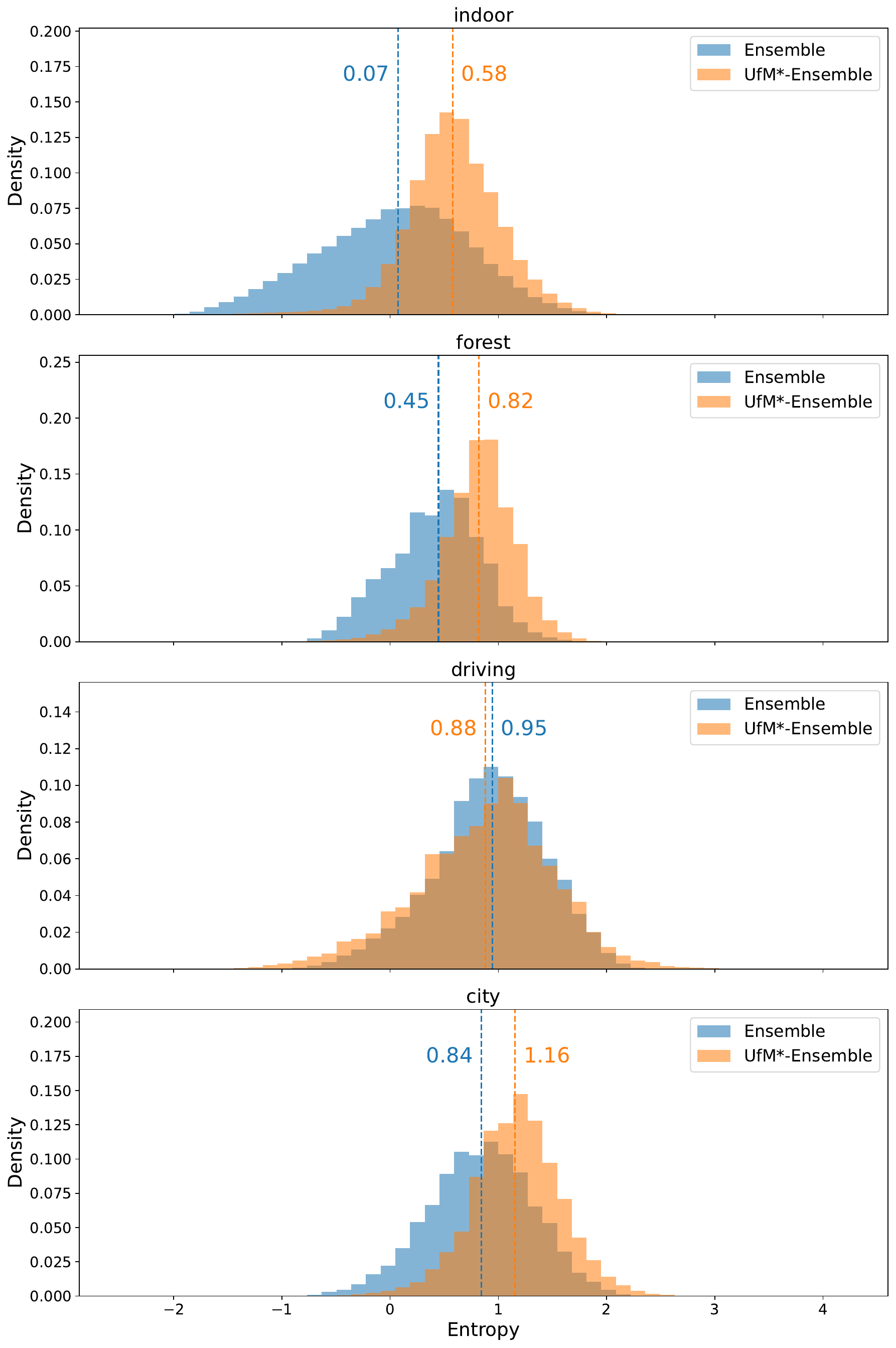}
\caption{Performance under distribution shift for ensemble and UfM$^*$-Ensemble methods, dashed lines show mean entropy.
}
\label{fig:entropy_by_category_histograms}
\end{figure}
\par We next evaluate UfM$^*$ on extremely out-of-distribution outdoor datasets where the depth DNN exhibits near-zero accuracy. In this setting, calibration metrics are not as informative, so instead we compare histograms of predictive entropy for the ensemble baseline and UfM$^*$-Ensemble. Fig.~\ref{fig:entropy_by_category_histograms} shows results for 100 ScanNet indoor sequences and for outdoor sequences from TartanAir and KITTI-360 spanning forest, driving, and city environments.
For the indoor sequences, the improved uncertainty quality of UfM$^*$-Ensemble arises from increasing uncertainty and reducing overconfidence relative to the baseline ensemble. However, for the highly out-of-distribution outdoor datasets, where prediction errors are extremely large, well-calibrated uncertainty would require substantially higher entropy. Instead, both the ensemble and UfM$^*$-Ensemble produce entropy values that remain comparatively low.

\par This behavior highlights a limitation of multiview disagreement: the predicted uncertainty is bounded by the range of depth predictions produced by the DNN. If the DNN predictions vary only within a small range (\textit{e.g.}, one meter) while the scale of the error is much higher (\textit{e.g.}, tens or hundreds of meters), multiview disagreement can detect only the limited disagreement among predictions, leading to overconfidence. Recent work on estimating scene scale for monocular depth prediction~\cite{bhat2023zoedepth, greene2020metrically} may help address this limitation by improving the scale of depth predictions, and we leave addressing this limitation for future work. 
\subsection{Ablations: pose noise, view overlap, no alternating} \label{sec:exp_results_pose_noise_overlap_robustness}

\begin{figure}[t!] \centering
         \centering
         \includegraphics[width=0.49\columnwidth]{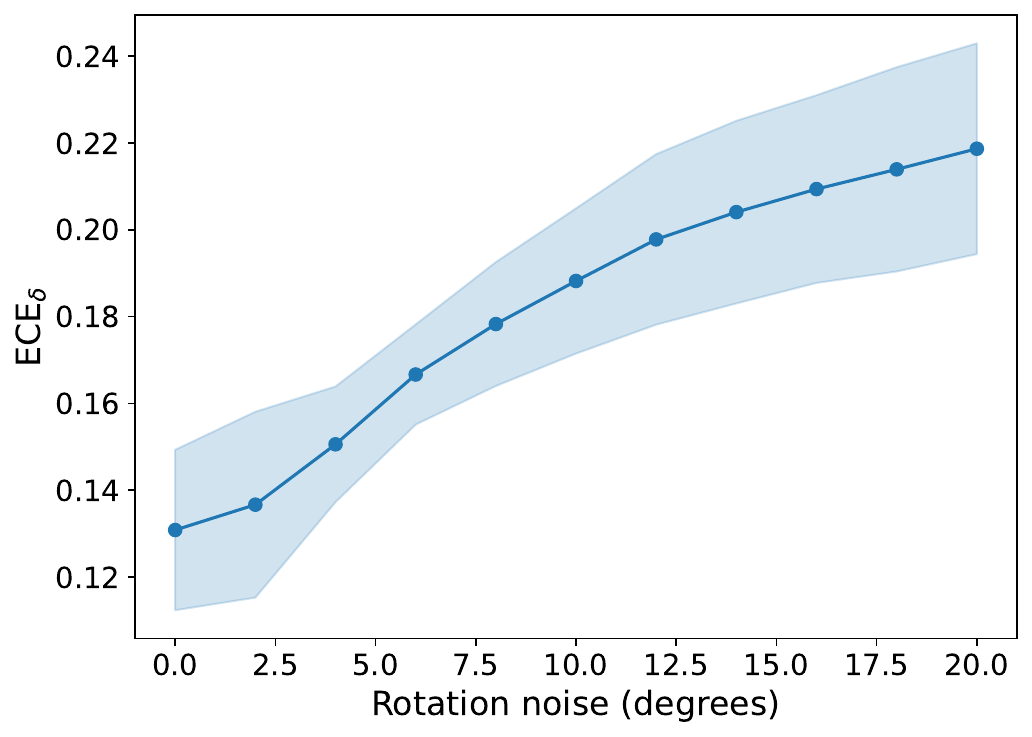}
         \includegraphics[width=0.49\columnwidth]{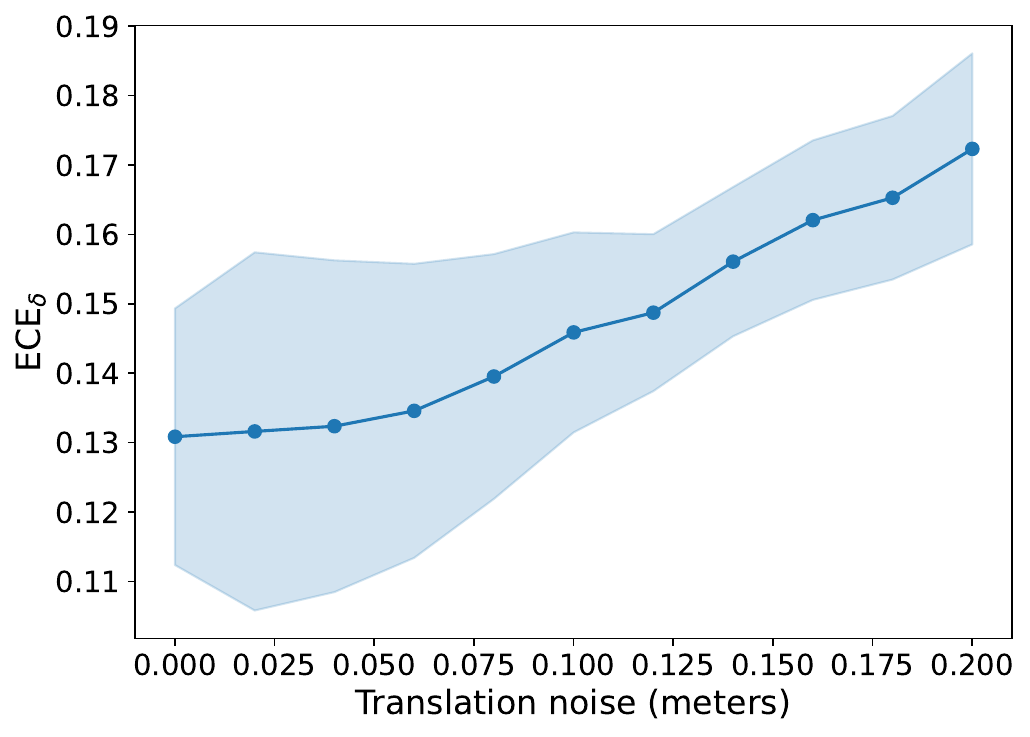}
     \caption{We show the robustness of UfM$^*$-Ensemble to rotation noise (left) and translation noise (right) added to the pose estimate from a SLAM system~\cite{dai2017bundlefusion}.}
     \label{fig:pose_noise}
\end{figure}

\begin{figure}[t!] \centering
         \centering
         \includegraphics[width=0.6\columnwidth]{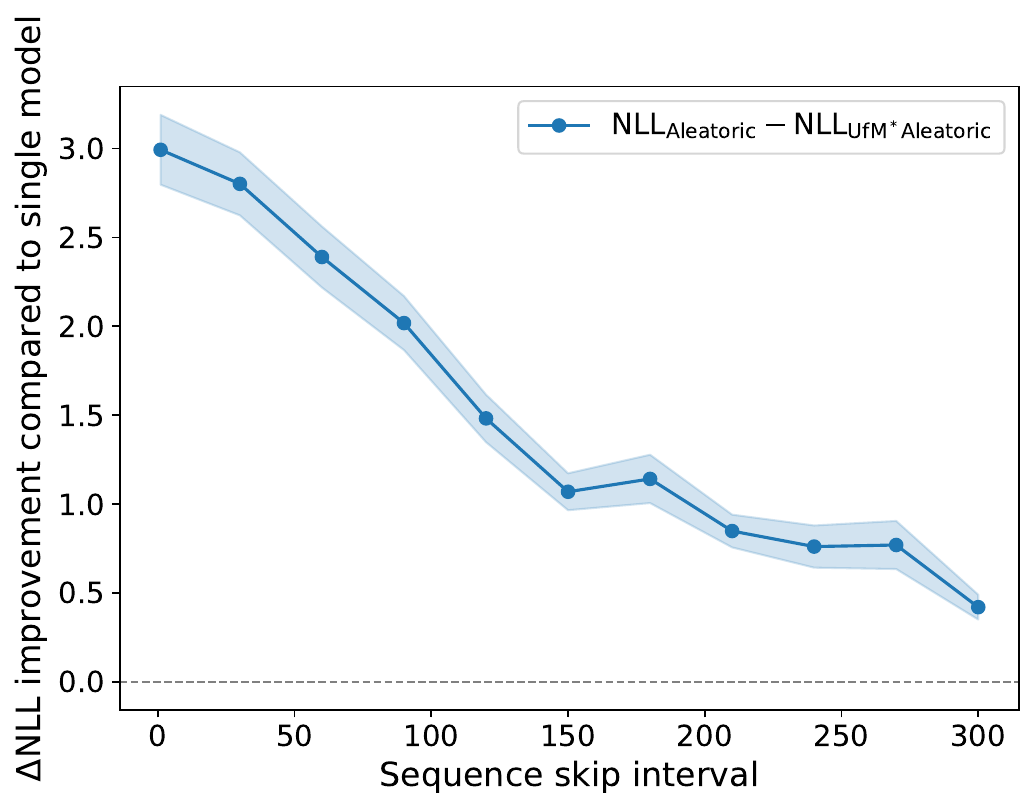}
     \caption{UfM$^*$-Aleatoric is robust to decreases in overlap between frames, still maintaining improvement over aleatoric uncertainty while converging towards it (gray dashed line). 
     }
     \label{fig:skip_interval}
\end{figure}

\begin{figure*}[t!] \centering
         \centering
         \includegraphics[width=1.0\textwidth]{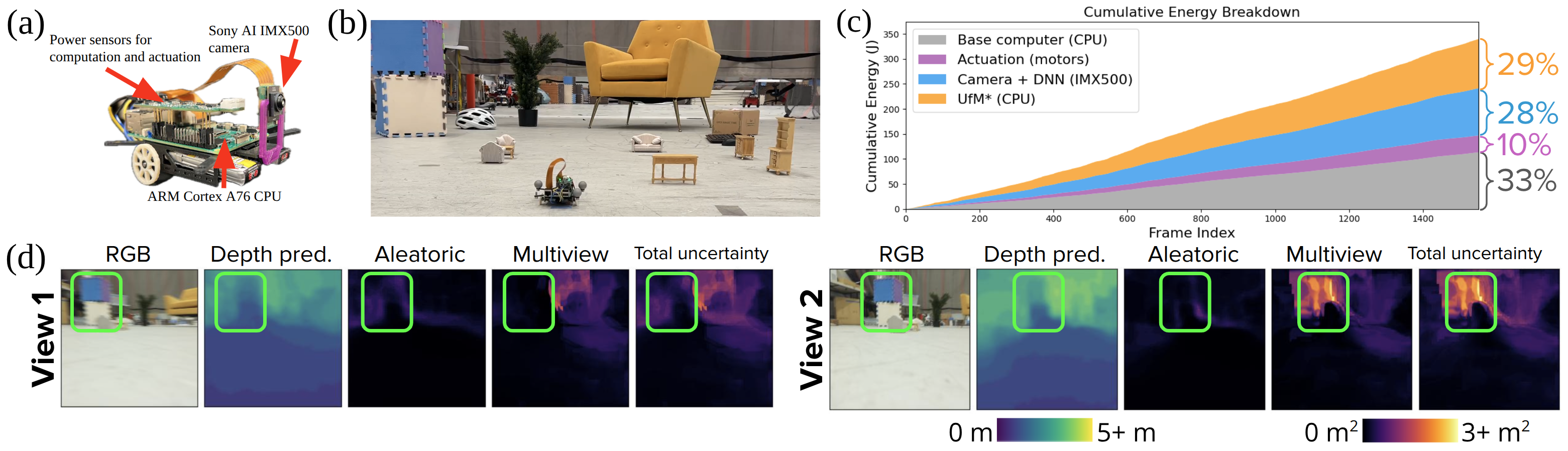}
     \caption{(a) Miniature car test platform with separate power sensors for computation and actuation, (b) experiment setup, (c) cumulative energy of uncertainty estimation vs. other tasks (camera + DNN, actuation, base computer power), (d) depth and uncertainty estimates on two views of a stack of cubes. UfM$^*$ is able to capture geometric inconsistency across views on the stack of cubes (boxed green) while running real-time at 30 FPS on an Arm Cortex-A76 CPU and consuming 63 mJ per image on average, where base computer power is measured as 2.22 W and IMX500 power is measured as 1.85 W.}
     \label{fig:system}
\end{figure*}
\begin{table}[t]
\centering
\caption{Ablation for multiple inference methods without alternating models using a ResNet50-Monodepth2 architecture on 100 ScanNet sequences.}
\label{tab:full_ufmstar_summary}

\setlength{\tabcolsep}{2pt}
\renewcommand{\arraystretch}{1.15}

{\footnotesize
\begin{tabularx}{\columnwidth}{%
    Y
    C{0.85cm} 
    C{0.95cm} 
    C{1.1cm} 
    C{1.1cm} 
    C{0.95cm} 
}
\toprule
Method & $\delta_1\uparrow$ 
& NLL $\downarrow$ 
& ECE$_{\delta}$ $\downarrow$ 
& ECE$_{\textit{q}}$ $\downarrow$ 
& Energy [mJ] \\
\midrule
UfM$^*$-FullEnsemble      & $\mathbf{0.70}$ & $\mathbf{0.72}$ & $\mathbf{0.12}$ & $0.15$ & $1810$ \\ 
UfM$^*$-FullMCD           & $0.68$          & $1.09$ & $0.13$ & $\mathbf{0.13}$ & $1815$ \\ 
UfM$^*$-FullBatchEnsemble & $0.68$          & $0.91$ & $0.12$ & $\mathbf{0.13}$ & $1478$ \\ 
\bottomrule
\end{tabularx}
}
\end{table}

We next present three ablations on the impact of pose noise, overlap between views, and multiple inferences. 
\par \textit{Pose noise:} Pose estimates are used in UfM$^*$ to (1) project Gaussians into 3D, (2) transform previously observed Gaussians into the current reference frame, and (3) project Gaussians into the 2D image plane to establish correspondences. Results on the ScanNet dataset, which provides poses estimated using BundleFusion~\cite{dai2017bundlefusion}, show that UfM$^*$ performs well using SLAM-derived poses and does not require ground-truth camera poses. To further examine robustness to pose errors, we inject synthetic noise into the camera rotation and translation and evaluate the resulting performance (Fig.~\ref{fig:pose_noise}). As expected, performance degrades as pose noise increases; however, UfM$^*$ maintains strong performance under modest levels of pose error.
\par \textit{Overlap between views:} Although our experiments use video sequences, UfM$^*$ does not require consecutive frames; it only requires sufficient overlap between views to observe the same 3D regions multiple times. To study the impact of view overlap, we vary the sequence skip interval in a ScanNet sequence and measure the resulting change in NLL. We present results in Fig.~\ref{fig:skip_interval}, where we plot the number of skipped images vs. the improvement in NLL achieved by UfM$^*$-Aleatoric relative to aleatoric baseline. When using consecutive frames (skip interval $=0$), UfM$^*$-Aleatoric provides the largest improvement in NLL compared to the aleatoric baseline. As the skip interval increases, the improvement gradually decreases and the performance of UfM$^*$-Aleatoric approaches that of the aleatoric baseline (gray dashed line). This behavior occurs because larger intervals between images reduce the overlap between views. Since multiview disagreement is zero in regions observed from only a single view, the total uncertainty increasingly relies on the single-view aleatoric uncertainty as overlap decreases.
\par \textit{Running UfM$^*$ with multiple inference methods:} UfM$^*$ does not require multiple inferences per image, unlike ensemble-based or sampling-based methods, because uncertainty is calculated across views while alternating models. Nevertheless, to evaluate the effect of combining multiview disagreement without alternating models, we run an ablation for UfM$^*$ with ensemble, MC-Dropout, and BatchEnsemble methods with multiple inferences per image  and report the results in Table~\ref{tab:full_ufmstar_summary}. We see that UfM$^*$-FullEnsemble achieves the best uncertainty quality. However, these approaches require multiple inferences per image which is very expensive.
\subsection{System demonstration: onboard multiview disagreement} \label{sec:exp_results_system_demo}
\par Finally, we demonstrate the impact of having a compute-efficient and memory-efficient uncertainty estimate that incorporates disagreement across views by deploying UfM$^*$ onboard a real-world miniature car test platform. Fig.~\ref{fig:system} illustrates the test platform, which carries a Raspberry Pi 5 computer with an Arm Cortex-A76 CPU and a Sony IMX500 AI camera accelerator capable of running compact DNN inference in real time onboard the platform. We quantize a FastDepth DNN, and deploy onboard the car in a motion capture lab with objects of varying scales (\textit{e.g.}, doll-sized to regular-sized furniture). UfM$^*$-Aleatoric estimates high multiview disagreement across views in regions where aleatoric uncertainty alone remains low (\textit{e.g.}, stack of cubes). It does so while remaining real-time on two CPU cores, showcasing DNN uncertainty estimation fully onboard a resource-constrained robot.
\section{Conclusion}
\label{sec:conclusion}
This work is enabled by three key insights. First, the presence of flickering depth predictions across views can be informative: since accurate depth is geometrically consistent, disagreement between predictions corresponding to the same 3D region provides a signal of uncertainty. Second, DNN depth uncertainty exhibits spatial structure; rather than treating pixels independently, uncertainty can be represented compactly in 3D using Gaussians that can better check regional inconsistency. Finally, because we seek to measure disagreement, we fuse measurements that should have agreed if accurate and quantify their actual disagreement instead of only fusing measurements that agree. Together, UfM$^*$ leverages multiview disagreement to make monocular depth DNN uncertainty estimation better calibrated and more efficient. 
\section{Acknowledgments}
The authors thank Noah Wiley for enabling perception onboard the miniature car test platform and helping with system experiments inside the motion capture laboratory.
\bibliographystyle{IEEEtran}
\bibliography{bib}
\end{document}